\documentclass[preprints,article,accept,oneauthor,pdftex]{Definitions/mdpi}

\firstpage{1} 
\pubvolume{xx}
\issuenum{1}
\articlenumber{5}
\pubyear{2020}
\copyrightyear{2020}
\history{Received: date; Accepted: date; Published: date}

\usepackage{csvsimple}
\usepackage{pdflscape}

\Title{Reduced Dilation-Erosion Perceptron for Binary Classification}

\Author{Marcos Eduardo Valle \orcidA{}}

\AuthorNames{Marcos Eduardo Valle}

\address[1]{%
Department of Applied Mathematics, University of Campinas; \\ E-mails: valle@ime.unicamp.br}

\corres{Correspondence: valle@ime.unicamp.br; Tel.: +55 19 3325-9375}

\newcommand{\R}{\mathbb{R}}

\newcommand{\vetx}{\boldsymbol{x}}
\newcommand{\vety}{\boldsymbol{y}}

\newcommand{\bb}{\begin{equation}}
\newcommand{\ee}{\end{equation}}
\newcommand{\bbb}{\begin{eqnarray}}
\newcommand{\eee}{\end{eqnarray}}
\newcommand{\benu}{\begin{enumerate}}
\newcommand{\eenu}{\end{enumerate}}

\newcommand{\vetw}{\boldsymbol{w}}
\newcommand{\vetm}{\boldsymbol{m}}

\newcommand{\bpm}{\begin{bmatrix}}
\newcommand{\epm}{\end{bmatrix}}

\newcommand{\sgn}{\text{sgn}}

\newcommand{\new}[1]{{#1}}

\abstract{Dilation and erosion are two elementary operations from mathematical morphology, a non-linear lattice computing methodology widely used for image processing and analysis. The dilation-erosion perceptron (DEP) is a morphological neural network obtained by a convex combination of a dilation and an erosion followed by the application of a hard-limiter function for binary classification tasks. A DEP classifier can be trained using a convex-concave procedure along with the minimization of the hinge loss function. As a lattice computing model, the DEP classifier assumes the feature and class spaces are partially ordered sets. In many practical situations, however, there is no natural ordering for the feature patterns. Using concepts from multi-valued mathematical morphology, this paper introduces the reduced dilation-erosion (r-DEP) classifier. An r-DEP classifier is obtained by endowing the feature space with an appropriate reduced ordering. Such reduced ordering can be determined using two approaches: One based on an ensemble of support vector classifiers (SVCs) with different kernels and the other based on a bagging of similar SVCs trained using different samples of the training set. Using several binary classification datasets from the OpenML repository, the ensemble and bagging r-DEP classifiers yielded in mean higher balanced accuracy scores than the linear, polynomial, and radial basis function (RBF) SVCs as well as their ensemble and a bagging of RBF SVCs.}


\begin{document}


\section{Introduction}

Cyber-physical systems (CPS) is a broad interdisciplinary area which combines  computational and physical devices in an integrated manner \cite{lee08,rajkumar10,shi11,derler12}. Internet of thinks (IoT), for instance, can be viewed as an important class of CPS where physical objects are interconnected in a network with identified address \cite{stojmenovic15}. Besides Industry 4.0 and related technologies \cite{oztemel20}, applications of CPS include social robots for educational purposes \cite{aspragathos19}, medical services and healthcare \cite{zhang17cps}, and many other fields.

Although modeling a CPS comprises both physical and computational process \cite{derler12}, this paper focuses only on the latter. Specifically, we model the computational process using lattice computing paradigm. Lattice computing (LC) comprises the many techniques and mathematical modeling methodologies based on lattice theory \cite{kaburlasos06,ritter07}. Lattice theory is concerned with a mathematical structure obtained by enriching a non-empty set with an ordering scheme with well defined extrema operations. Specifically, a lattice is a partially ordered set in which any finite set has both an infimum and a supremum \cite{birkhoff93}. 
One of the main advantages of LC is its capability to process ordered data which include logic values, sets and more generally fuzzy sets, images, graphs, and many other types of information granules \cite{kaburlasos06,ritter07,skowron01,yao13}. Mathematical morphology and morphological neural networks are examples of well succeeded LC modeling methodologies. Let us briefly address these two LC methodologies in the following paragraphs.  

Mathematical morphology (MM) is a non-linear theory widely used for image processing and analysis \cite{serra82,serra88,soille99,najman13}. MM has been originally conceived for processing binary images in the 1960s. Subsequently, it has been extended to gray-scale images using the notions of umbra, level sets, and fuzzy set theory \cite{sternberg86,nachtegael01,sussner08jmiv}. Complete lattice is one key concept to extend MM from binary to more general contexts \cite{ronse90,nachtegael11}. Specifically, MM is a theory mainly concerned with mappings between complete lattices \cite{heijmans94,heijmans95}. Dilations and erosions, which are defined algebraically as mappings that commute respectively with the supremum and infimum operations, are elementary operations of MM. Many other operators from MM are defined by combining dilations and erosions \cite{soille93,najman13,heijmans94}. 
Although complete lattice provide an appropriate mathematical background for binary and gray-scale MMs, defining morphological operators for multi-valued images is not straightforward because there is no universal ordering for vector-valued spaces \cite{angulo07,aptoula07}. In fact, the development of appropriate ordering scheme for multi-valued MM is an active area of research \cite{burgeth14,burgeth19ismm,lezoray16,sangalli19ismm,velasco-forero14}. In this paper, we make use of the suprevised reduced ordering proposed by Velasco-Forero and Angulo \cite{velasco-forero11a}. In few words, a supervised reduced ordering is defined using a training set of negative (or background) and positive (or foreground) values. As a consequence, the resulting multi-valued morphological operators can be interpreted in terms of positive and negative training values. One contribution of this paper is the definition of morphological neural networks based on supervised reduced orderings (see Section \ref{sec:r-DEP}). 

Morphological neural networks (MNN) refer to the broad class of neural networks whose processing units perform an operation from MM, possibly followed by the application of an activation function \cite{sussner11ins}.  The single-layer moprhological perceptron, introduced by Ritter and Sussner in the middle 1990s, is one of the earliest MNNs along with the morphological associative memories \cite{ritter96c}. Briefly, the morphological perceptron performs either a dilation or an erosion from gray-scale MM followed by a hard-limiter activation function. The original morphological perceptron has been subsequently investigated and generalized by many prominent researchers \cite{sussner98,ritter03,sussner09ijcnn,sussner11ins}. For example, Sussner addressed the multilayer morphological perceptron and introduced a supervised learning algorithm for binary classification problems \cite{sussner98}. In few words, the learning algorithm proposed by Sussner is an incremental algorithm which adds hidden morphological neurons until all the training set is correctly classified. By taking into account the relevance of dendrites in biological neurons, Ritter and Urcid proposed a morphological neuron with dendritic structure \cite{ritter03}. Apart from the biological motivation, the morphological perceptron with dendritic structure is similar to the multilayer morphological perceptron investigated by Sussner. In fact, like the learning algorithm of the multilayer morphological perceptron, the morphological perceptron with dendritic structure grows as it learns until there is no miss-classified training samples. Furthermore, the decision surface of both multilayer perceptron and the morphological perceptron with dentritic structure depends on the order in which the training samples are presented to the network. The morphological perceptron with competitive layer (MPC) introduced by Sussner and Esmi does not depend on the order in which the training samples are presented to the network \cite{sussner11ins}. Like the previous models, however, training morphological perceptron with competitive layer finishes only when all training patterns are correctly classified. Thus, it is possible to the network to end up overfitting the training data.

In contrast to the greedy methods described in the previous paragraph, many researchers formulated the training of MNNs and hybrid models as an optimization problem. For example, Pessoal and Maragos used pulse functions to circumvent the non-differentiability of lattice-based operations on a steepest descent method designed for training a hybrid morphological/rank/linear network \cite{pessoa00}. Based on the ideas of Pessoa and Maragos, Araujo proposed a hybrid morphological/linear network called {dilation-erosion perceptron} (DEP), which is trained using a steepest descent method \cite{araujo11}. Steepest descent methods are also used by Hern{\'a}ndez et al. for training hybrid two-layer neural networks, where one layer is morphological and the other is linear  \cite{hernandez19}. \new{In a similar fashion, Mondal et al. proposed an hybrid morphological/linear model, called dense morphological network, which is trained using stochastic gradient descent method such as adam optimization \cite{mondal19dense}.}
Apparently unaware of the aforementioned works on MNNs, Franchi et al. recently integrated morphological operators with a deep learning framework to introduce the so-called deep morphological network, which is also trained using a steepest descent algorithm \cite{franchi20}. In contrast to steepest descent methods, Arce et al. trained MNNs using differential evolution \cite{arce18}. Moreover, Sussner and Campiotti proposed a hybrid morphological/linear extreme learning machine which has a hidden-layer of morphological units and  a linear output layer that is trained by regularized least-squares \cite{sussner20nn}. 
Recently, Charisopoulos and Maragos formulated the training of a single morphological perceptron as the solution of a convex-concave optimization problem \cite{charisopoulos17,shen16}. Apart from the elegant formulation, the convex-concave procedure outperformed gradient descent methods in terms of accuracy and robustness on some computational experiments. 

In this paper, we investigate the convex-concave procedure for training the DEP classifier. As a lattice-based model, DEP requires a partial ordering on both feature and class label spaces. Furthermore, the traditional approach assumes the feature space is equipped with the component-wise ordering induced by the natural ordering of real numbers. The component-wise ordering, however, may be inappropriate in the feature space. Based on ideas from multi-valued MM, we make use of supervised reduced orderings for the feature space of the DEP model. The resulting model is referred to as reduced dilation-erosion perceptron (r-DEP). The performance of the new r-DEP model is evaluated by considering 30 binary classification problems from the OpenML repository \cite{OpenML,OpenMLPython}, in which most of them are also available at the well-known UCI Machine Learning Repository \cite{uci_repository}.

The paper is organized as follows. The next section presents a brief review on the basic concepts from lattice theory and MM, including the supervised reduced ordering-based approach to multi-valued MM. Traditional MNNs, including the DEP classifier and the convex-concave procedure, are discussed in Section \ref{sec:MNN}. Section \ref{sec:r-DEP} presents the main contribution of this paper: the reduced DEP classifier. In Section \ref{sec:Experiments}, we compare the performance of the r-DEP classifier with other traditional machine learning approaches from the literature. The paper finishes with some concluding remarks in Section \ref{sec:Concluding}.


\section{Basic Concepts from Lattice Theory and Mathematical Morphology}

Let us begin by recalling some basic concepts from lattice theory and mathematical morphology (MM). Precisely, we shall only present the necessary concepts for understanding of morphological perceptron models. Furthermore, we will focus on elementary concepts without going deep into these rich theories. The reader interested on lattice theory is invited to consult \cite{birkhoff93}. Detailed account on MM and its applications can be found in \cite{serra82,soille93,najman13,heijmans94}. The reader familiar with lattice theory and MM may skip subsection \ref{ssec:Lattice-Theory}. 

\subsection{Lattice Theory and Mathematical Morphology} \label{ssec:Lattice-Theory}

First of all, a non-empty set $\mathbb{L}$ equipped with a binary relation "$\leq$" is a {partially ordered set} (poset) if the following conditions hold true:

\begin{enumerate}
\item[P1:] For all $x \in \mathbb{L}$, we have $x \leq x$. \hfill (Reflexive)
\item[P2:] If $x \leq y$ and $y \leq z$, then $x \leq z$. \hfill (Transitivity)
\item[P3:] If $x \leq y$ and $y \leq x$, then $x = y$. \hfill (Antisymmetry)
\end{enumerate}
In this case, the binary relation ``$\leq$'' is called a partial order. We speak of a {pre-ordered} set if $\mathbb{L}$ is equipped with a binary relation which satisfies the properties P1 and P2.

A partially ordered set $\mathbb{L}$ is a {complete lattice} if any subset $X \subset \mathbb{L}$ has a supremum (least upper bound) and an infimum (greatest lower bound) denoted respectively by $\bigvee X$ and $\bigwedge X$. When $X = \{x_1,\ldots,x_n\}$ is finite, we write $\bigwedge X = \bigwedge_{i=1}^n$ and $\bigvee X = \bigvee_{i=1}^n x_i$. 

\begin{Example}
The extended real numbers $\bar{\mathbb{R}} = \mathbb{R} \cup \{+\infty,-\infty\}$ with the natural ordering is an example of a complete lattice. The Cartesian product of the extended real numbers $\bar{\mathbb{R}}^n$ is also a complete lattice with the partial order defined as follows in a component-wise manner:
\begin{equation} \label{eq:marginal}
    \boldsymbol{x}=(x_1,x_2,\ldots,x_n) \leq (y_1,y_2,\ldots,y_n) = \boldsymbol{y}  \quad \Longleftrightarrow \quad x_i \leq y_i, \forall i=1,\ldots,n.
\end{equation}
In this case, the infimum and the supremum of a set $X \subseteq \bar{\R}^n$ is also determined in a component-wise manner by
\begin{equation}
    \bigwedge X = \left( \bigwedge X_1, \bigwedge X_2,\ldots, \bigwedge X_n\right) \quad \mbox{and} \quad 
    \bigvee X = \left( \bigvee X_1, \bigvee X_2,\ldots, \bigvee X_n\right),
\end{equation}
where $X_i = \{ x_i: (x_1,\ldots,x_i,\ldots,x_n) \in X \} \subseteq \bar{\R}$ is the set of the $i$th component of all the vectors in $X$, for $i=1,\ldots,n$.
\end{Example}

Mathematical morphology is a non-linear theory widely used for image processing and analysis \cite{soille93,heijmans95}. From the mathematical point of view, MM can be viewed as a theory of mappings between complete lattices. In fact, the elementary operations of MM are mappings that distribute over either infimum or supremum operations. Precisely, two elementary operations of MM are defined as follows \cite{serra88,heijmans94}: 

\begin{Definition}[Erosion and Dilation] \label{def:MM}
Let $\mathbb{L}$ and $\mathbb{M}$ be complete lattices. A mapping $\varepsilon: \mathbb{L} \to \mathbb{M}$ is an {erosion} and a mapping $\delta:\mathbb{L} \to \mathbb{M}$ is a {dilation} if the following identities hold true for any $X \subseteq \mathbb{L}$ :
\begin{equation} \label{eq:ero_dil}
 \varepsilon \left(\bigwedge X\right) = \bigwedge_{x \in X} \varepsilon \left(x \right) \quad \mbox{and} \quad \delta \left(\bigvee X \right) = \bigvee_{x \in X} \delta \left(x \right).    
\end{equation}
\end{Definition}

From Theorem 3 on \cite{sussner11ins}, we have the following example of dilations and erosions:
\begin{Example}
Consider real-valued vectors $\vetm = (m_1,\dots,m_n) \in \R^n$ and $\vetw = (w_1,\ldots,w_n) \in \R^n$. The operators $\varepsilon_{\vetm}: \bar{\R}^n \to \bar{\R}$ and $\delta_{\vetw}: \bar{\R}^n \to \bar{\R}$ given by 
\begin{equation} \label{eq:MMw}
\varepsilon_{\vetm}(\vetx) = \bigwedge_{j=1}^n (m_j + x_j) \quad \mbox{and} \quad \delta_{\vetw}(\vetx) = \bigvee_{j=1}^n (w_j + x_j),
\end{equation}
for all $\vetx = (x_1,x_2,\ldots,x_n) \in \bar{\R}^n$ are respectively an erosion and a dilation. 
\end{Example}

\begin{Remark}
The mappings $\varepsilon_{\vetm}$ and $\delta_{\vetw}$ given by \eqref{eq:MMw} can be extended for $\vetm,\vetw \in \bar{\R}^n$ by appropriately dealing with indeterminacy such as $(+\infty)+(-\infty)$ \cite{sussner11ins}. In this paper, however, we only consider finite-valued vectors $\vetm,\vetw \in \R^n$.
\end{Remark}

\begin{Definition}[Increasing Operators] \label{def:Increasing}
Let $\mathbb{L}$ and $\mathbb{M}$ be two complete lattices. An operator $\psi: \mathbb{L} \to \mathbb{M}$ is isotone or increasing if $x \leq y$ implies \new{$\psi(x) \leq \psi(y)$}. 
\end{Definition}

\begin{Proposition}[Lemma 2.1 from \cite{heijmans90}]
Erosions and dilations are increasing operators. 
\end{Proposition}

Despite the rich theory on morphological operators and their many successful applications, the concepts present above are sufficient for this paper. Let us now turn our attention to some concepts from multi-valued MM. 

\subsection{Multi-valued Mathematical Morphology}

Although MM can be very well defined on complete lattices (see Definition \ref{def:MM}), there is no unambiguous ordering for vector-valued sets. For example, although $\bar{\R}^n$ equipped with the component-wise ordering is a complete lattice, the partial order given by \eqref{eq:marginal} does not take into account possible relationship between the vector components. Furthermore, the component-wise order given by \eqref{eq:marginal} results the so-called ``false color'' problem in multi-valued MM \cite{serra09}.
As a consequence, a great deal of effort has been devoted to finding appropriate ordering schemes for vector-valued data \cite{goutsias95,angulo07,aptoula07,sangalli19ismm,burgeth19ismm}. Among the many approaches to multi-valued MM, those based on reduced orderings are particularly interesting and computationally cheap \cite{goutsias95,velasco-forero11a,velasco-forero14}.

In a reduced ordering, also referred to as an r-ordering, the elements of a vector-valued non-empty set $\mathbb{V}$ are ranked according to a surjective mapping $\rho:\mathbb{V} \to \mathbb{L}$, where $\mathbb{L}$ is a complete lattice. Precisely, an r-ordering is defined as follows using the mapping $\rho:\mathbb{V} \to \mathbb{L}$:
\begin{equation} \label{eq:r-ordering} \vetx \leq_\rho \vety \quad \Longleftrightarrow \quad \rho(\vetx) \leq \rho(\vety), \quad \forall \vetx,\vety \in \mathbb{V}.\end{equation}

In analogy to Definition \ref{def:Increasing}, r-increasing operators are defined as follows using r-orderings:
\begin{Definition}[r-Increasing Operator] \label{def:r-increasing} 
Let $\rho:\mathbb{V} \to \mathbb{L}$ and $\sigma:\mathbb{W} \to \mathbb{M}$ be surjective mappings from non-empty sets $\mathbb{V}$ and $\mathbb{W}$ to complete lattices $\mathbb{L}$ and $\mathbb{M}$. An operator $\psi:\mathbb{V} \to \mathbb{W}$ is r-increasing if $\vetx \leq_{\rho} \vety$ implies $\psi(\vetx) \leq_{\sigma} \psi(\vety)$.
\end{Definition}


Although been reflexive and transitive, an r-ordering is in principle a pre-ordering because it may fails to be anti-symmetric. Notwithstanding, morphological operators can be defined as follows using reduced orderings \cite{goutsias95}:
\begin{Definition}[r-Increasing Morphological Operator] \label{def:r-MM}
Let $\mathbb{V}$ and $\mathbb{W}$ be non-empty sets, $\mathbb{L}$ and $\mathbb{M}$ be complete lattices, and $\rho:\mathbb{V} \to \mathbb{L}$ and $\sigma:\mathbb{W} \to \mathbb{M}$ be surjective mappings. A mapping $\psi^{r}:\mathbb{V} \to \mathbb{W}$ is an r-increasing morphological operator if there exists an increasing morphological operator $\psi:\mathbb{L} \to \mathbb{M}$ such that $\sigma \psi^{r} = \psi \rho$, that is, 
\begin{equation}
    \sigma \big( \psi^{r}(\vetx) \big) = \psi \big( \rho(\vetx) \big), \quad \forall \vetx \in \mathbb{V}.
\end{equation}
\end{Definition}

In words, the mapping $\sigma:\mathbb{W} \to \mathbb{M}$ applied on an r-increasing morphological operator $\psi^{r}:\mathbb{V} \to \mathbb{W}$ corresponds to the morphological operator $\psi:\mathbb{L} \to \mathbb{M}$ applied on the output of the mapping $\rho:\mathbb{V} \to \mathbb{L}$. 
For example, an operator $\varepsilon^r:\mathbb{V} \to \mathbb{W}$ is an r-increasing erosion, or simply an r-erosion, if there exists an erosion $\varepsilon:\mathbb{L} \to \mathbb{M}$ such that $\sigma \varepsilon^r = \varepsilon \rho$. Dually, a mapping $\delta^{r}:\mathbb{V} \to \mathbb{W}$ is an r-increasing dilation, or simply an r-dilation, if there exists a dilation $\delta:\mathbb{L} \to \mathbb{M}$ such that $\sigma \delta^{r} = \delta \rho$.

\begin{Remark}
Definition \ref{def:r-MM}, which is motivated by Proposition 2.5 in \cite{goutsias95}, generalizes of the notions of r-erosion and r-dilation as well as r-opening and r-closing from Goutsias et al. Precisely, if $\mathbb{W}=\mathbb{V}$ and $\mathbb{M}=\mathbb{L}$, then one can consider $\sigma = \rho$, where $\rho:\mathbb{V} \to \mathbb{L}$ is a surjective mapping. Thus, an operator $\varepsilon^r:\mathbb{V} \to \mathbb{V}$ is an r-erosion if and only if there exists an erosion $\varepsilon:\mathbb{L} \to \mathbb{L}$ such that $\rho \varepsilon^r = \varepsilon \rho$, which is exactly the definition of r-erosion introduced by Goutsias et al. (shortly after Proposition 2.11 in \cite{goutsias95}). 
\end{Remark}


\begin{Example}
A general approach to multi-valued MM based on supppervised reduced ordering have been proposed by Velasco-Forero and Angulo \cite{velasco-forero11a}. Briefly, in a supervised reduced ordering the sobrejective mapping $\rho:\mathbb{V} \to \mathbb{L}$ is determined using training sets $\mathcal{P}$ and $\mathcal{N}$ of positive (foreground) and negative (background) values, respectively. Furthermore, the mapping $\rho$ is expected to satisfy 
\begin{equation}
    \rho(\vetx) = \begin{cases}
    \top, & \vetx \in \mathcal{P},\\
    \bot, & \vetx \in \mathcal{N},
    \end{cases} 
\end{equation}
where $\top = \bigvee \mathbb{L}$ and $\bot = \bigwedge \mathbb{L}$ denote respectively the largest (top) and the least (bottom) elements of $\mathbb{L}$. As a consequence, a supervised r-ordering is interpretable with respect to the training sets $\mathcal{P}$ and $\mathcal{N}$ \cite{velasco-forero14}. Assuming $\mathbb{V} \subset \R^N$ and $\mathbb{L} \subseteq \bar{\R}$, Velasco-Forero and Angulo proposed to determine $\rho$ using a support vector machine \cite{vapnik98,smola02,haykin09}. As usual, let us first combine the positive and negative sets into a single training set $\mathcal{T} = \{(\vetx_i,d_i):i=1,\ldots,M\}$ such that $d_i=+1$ if $\vetx_i \in \mathcal{P}$ and $d_i = -1$ if $\vetx_i \in \mathcal{N}$. Given a suitable kernel function $\kappa$, the supervised reduced ordering mapping $\rho$ corresponds to the decision function of a support vector classifier (SVC) given by
\begin{equation} \label{eq:r-SVM} 
\rho(\vetx) = \sum_{i=1}^M \alpha_i d_i \kappa(\vetx,\vetx_i),
\end{equation}
where $\boldsymbol{\alpha} = (\alpha_1,\ldots,\alpha_M) \in \R^M$ is the solution of the quadratic programming problem
\begin{align}
    \label{eq:SVM-QP}
    &\mathop{\mbox{minimize}}_{\boldsymbol{\alpha}} \quad Q(\boldsymbol{\alpha}) =  \frac{1}{2} \sum_{i,j=1}^M \alpha_i \alpha_j d_i d_j \kappa(\vetx_i,\vetx_j) - \sum_{i=1}^M \alpha_i, \\
    &\mbox{subject to} \quad \sum_{i=1}^M \alpha_i d_i = 0 \quad \mbox{and} \quad 0 \leq \alpha_i \leq C, \quad \forall i=1,\ldots,M,
\end{align}
where $C > 0$ is a user specified parameter which controls the trade-off between minimizing the training error and maximizing the separation margin \cite{smola02,haykin09}. Examples of kernels include
\begin{itemize}
    \item Linear kernel: 
    \begin{equation} \label{eq:kernel-linear} \kappa(\vetx,\vety) = \langle \vetx,\vety \rangle. 
    \end{equation}
    \item Gaussian kernel: 
    \begin{equation} \label{eq:kernel-RBF} \kappa(\vetx,\vety) = e^{-\|\vetx-\vety\|^2/(2\sigma^2)}. \end{equation}
    The Gaussian kernel yields a radial-basis function (RBF) SVC.
    \item Polynomial kernel: 
    \begin{equation} \label{eq:kernel-poly} \kappa(\vetx,\vety) = \langle \vetx,\vety \rangle^d. \end{equation}
    The polynomial kernel yields a polynomial SVC.
\end{itemize}
We would like to point out that the intercept term is not relevant for a reduced ordering scheme and, thus, we refrained to include it on \eqref{eq:r-SVM}. Also, we would like to remark that the decision function $\rho$ maps the multi-valued set $\mathbb{V}$ to a totally ordered subset of $\bar{\mathbb{R}}$, which allows for efficient implementation of multi-valued morphological operators using look-up table and the usual gray-scale operators (see Algorithm 1 in \cite{velasco-forero14} for details).
\end{Example}

\section{Morphological Perceptron and the Convex-Concave Procedure} \label{sec:MNN}

Morphological neural network (MNN) refer to the broad class of neural networks whose neurons (processing units) perform an operation from MM possibly followed by the application of an activation function \cite{sussner11ins}. Examples of MNNs include (fuzzy) morphological associative memories \cite{ritter98,ritter99,sussner06nn,valle08fss,valle11nn,santos18nn} and morphological perceptrons \cite{ritter96c,sussner98,sussner09ijcnn,sussner11ins,charisopoulos17}. In this paper, we focus on the dilation-erosion perceptron trained using the recent convex-concave procedure and applied as a binary classifier \cite{araujo11,charisopoulos17,shen16}. 

Recall that a classifier is a mapping $\phi:\mathbb{V} \to \mathcal{C}$, where $\mathbb{V}$ and $\mathcal{C}$ are respectively the sets of features and classes. In a binary classification problem, the set $\mathcal{C}=\{c_1,c_2\}$ of classes can be identified with $\{-1,+1\}$ by means of a one-to-one mapping $\sigma:\mathcal{C} \to \{-1,+1\}$. 
\new{In this paper, we say that the elements of $\mathbb{V}$ associated with the class labels $-1$ and $+1$ belong respectively to the negative and positive classes.} In a supervised binary classification task, the classifier $\phi:\mathbb{V} \to \{-1,+1\}$ is determined from a finite set of samples $\mathcal{T} = \{(\vetx_i,d_i):i=1,\ldots,m\} \subset \mathbb{V} \times \{-1,+1\}$ referred to as the training set. 

\subsection{Morphological and Dilation-Erosion Perceptron Models}

Morphological perceptron has been introduced by Ritter and Sussner in the middle 1990s for binary classification problems \cite{ritter96c}. In analogy to the Rosenblatt's perceptron, Ritter and Sussner define a morphological perceptron by either one of the two equations 
\begin{equation} \label{eq:morph-perceptron}
    y = f \left( \bigwedge_{j=1}^n (m_j+x_j) \right) \quad \mbox{or} \quad y = f \left( \bigvee_{j=1}^n (w_j + x_j) \right),
\end{equation}
where $f$ denotes a hard limiter activation function. To simplify the exposition, we consider in this paper $f \equiv \sgn$ with the convention $\sgn(0)=+1$. By adopting the signal function, the morphological perceptrons given by \eqref{eq:morph-perceptron} can be used for binary classification whose labels are $-1$ and $+1$.

Note that a morphological perceptron is given by either the composition $f \varepsilon_{\vetm}$ or the composition $f \delta_{\vetw}$, where $\varepsilon_{\vetm}:\bar{\R}^n \to \bar{\R}$ and $\delta_{\vetw}:\bar{\R}^n \to \bar{\R}$ denote respectively the dilation and the erosion given by \eqref{eq:MMw} for $\vetm, \vetw \in \R^n$. Therefore, we refer to the models in \eqref{eq:morph-perceptron} as erosion-based and dilation-based morphological perceptrons, respectively. Furthermore, $\varepsilon_{\vetm}:\bar{\R}^n \to \bar{\R}$ and $\delta_{\vetw}:\bar{\R}^n \to \bar{\R}$ are respectively the decision functions of the erosion-based and dilation-based morphological perceptron.

\begin{figure}
    \centering
    \includegraphics[width = 0.48\columnwidth]{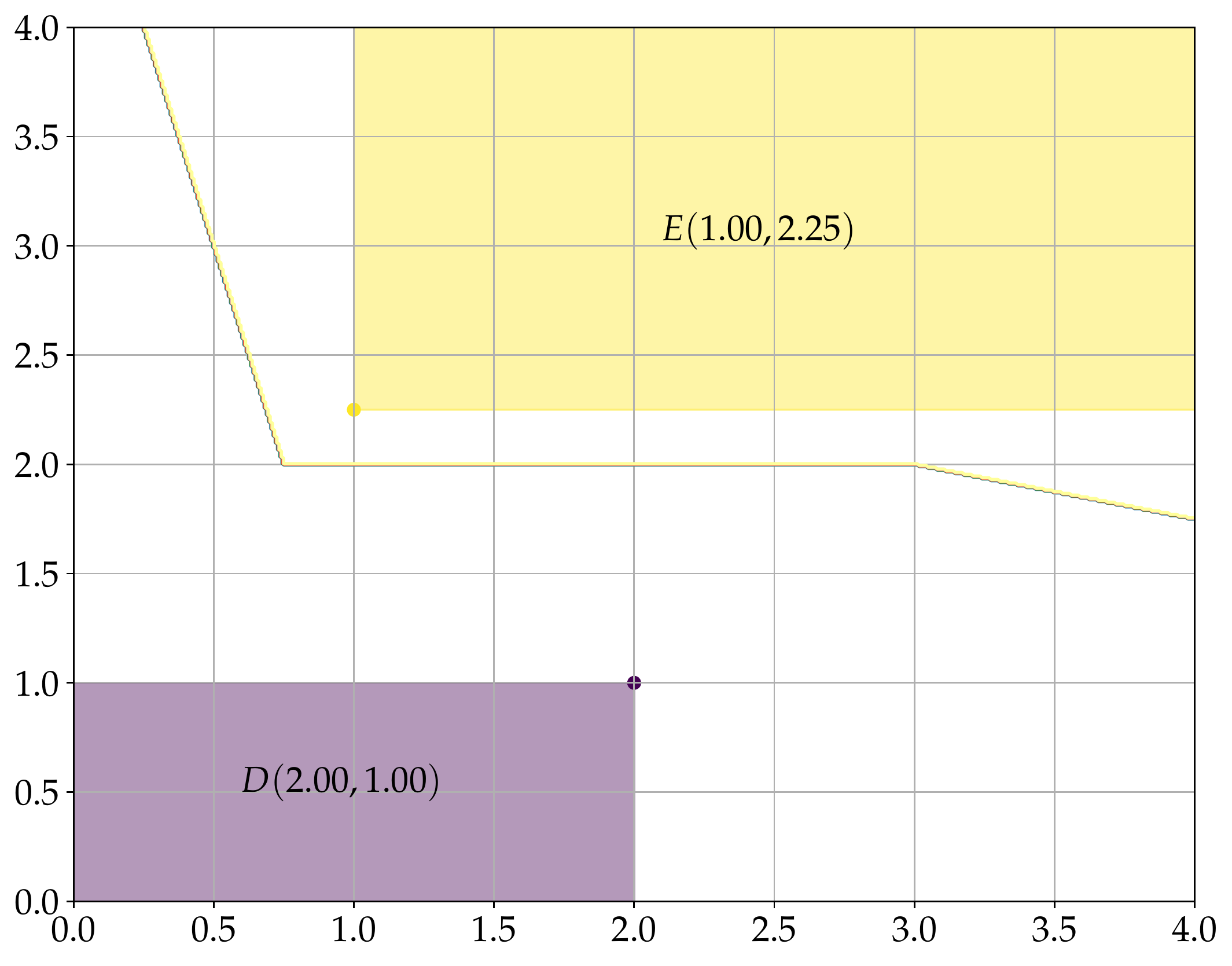}
    \caption{\new{The purple and the yellow regions corresponds respectively to the sets $E(1,2.25)$ and $D(2,1)$. The piece-wise linear curve corresponds to the decision boundary of the DEP classifier with $\beta = 0.2$.}}
    \label{fig:Example1}
\end{figure}

Let us now briefly address the geometry of the morphological perceptrons with $f \equiv \sgn$. Given a weight vector $\vetm = (m_1,\ldots,m_n) \in \R^n$, let 
\begin{equation}
    E(\vetm) = \{\vetx \in \R^n: \varepsilon_{\vetm}(\vetx) \geq 0 \},
\end{equation}
be the set of all points such that $y = \sgn \varepsilon_{\vetm}(\vetx) \geq 0$. Since $\varepsilon_{\vetm}(\vetx) = \bigwedge_{j=1}^n (m_j + x_j) \geq 0$ if and only if $m_j + x_j \geq 0$ for all $j=1,\ldots,n$, we conclude that $E(\vetm)$ is equivalently given by 
\begin{equation} \label{eq:region-E}
    E(\vetm) = \{\vetx = (x_1,\ldots,x_n) \in \R^n: x_j \geq -m_j, \forall j=1,\ldots,n\}.
\end{equation}
The decision boundary of an erosion-based morphological perceptron $\sgn \varepsilon_{\vetm}$ corresponds to the boundary of the set $E(\vetm)$: The class label $+1$ is assigned to all patterns in $E(\vetm)$ while the class label $-1$ is given to all patterns outside $E(\vetm)$. In view of this remark, we may say that an erosion-based morphological perceptron focuses on the positive class, whose label is $+1$. Dually, a dilation-based morphological perceptron focuses on the negative class, whose label is $-1$. Specifically,   
given a weight vector $\vetw=(w_1,\ldots,w_n) \in \R^n$, the set $D(\vetw)$ of all points $\vetx \in \R^n$ such that $y = \sgn \delta_{\vetw}(\vetx) < 0$ satisfies
\begin{equation} \label{eq:region-D}
    D(\vetw) = \{\vetx = (x_1,\ldots,x_n) \in \R^n: x_j < -w_j, \forall j=1,\ldots,n\}.
\end{equation}
The decision boundary of a dilation-based morphological perceptron corresponds to the boundary of $D(\vetw)$, that is, patterns inside $D(\vetw)$ are classified as negative while patterns outside $D(\vetw)$ are classified as $+1$. 
For illustrative purposes, Figure \ref{fig:Example1} shows the sets $E(1,2.25)$ (yellow region) and $D(2,1)$ (purple region), obtained by considering respectively $\vetw = (2,1) \in \R^2$ and $\vetm=(1,2.25) \in \R^2$. Figure \ref{fig:Example1} also shows the decision boundary of the dilation-erosion perceptron described below. 

In the previous paragraph, we pointed out that erosion-based and dilation-based morphological perceptrons focus respectively on the positive and the negative classes. The dilation-erosion perceptron (DEP) proposed by Ara\'ujo allows a graceful balance between the two classes \cite{araujo11}. The dilation-erosion perceptron is simply a convex combination of an erosion-based and a dilation-based morphological perceptron. In mathematical terms, given $\vetm,\vetw \in \R^n$ and
$0 \leq \beta \leq 1$, the decision function of a DEP classifier is defined by 
\begin{equation} \label{eq:DEP}
  \tau(\vetx) = \beta \delta_{\vetw}(\vetx) + (1-\beta) \varepsilon_{\vetm}(\vetx), \quad \forall \vetx \in \R^n.
\end{equation}
The binary DEP classifier $\phi:\R^n \to \{-1,+1\}$ is defined by the composition 
\begin{equation} \label{eq:DEP-classifier}
\phi = \sgn \tau.    
\end{equation}
In other words, given $\vetm,\vetw \in \R^n$ and $\beta \in [0,1]$, the class of an unknown pattern $\vetx \in \R^n$ is determined by evaluating $\phi(\vetx) = \sgn \tau(\vetx)$.

Note that $\tau$ given by \eqref{eq:DEP} corresponds respectively to $\delta_{\vetw}$ and $\varepsilon_{\vetm}$ when $\beta=0$ and $\beta=1$. More generally, the parameter $\beta$ controls the trade off between the dilation-based and the erosion-based morphological perceptrons, which focus on negative and positive classes, respectively.  For illustrative purposes, the decision boundary of a DEP classifier obtained by considering $\beta=0.2$, $\vetm=(1,2.25)$ and $\vetw = (2,1)$ is depicted in Figure \ref{fig:Example1}. In this case, the decision boundary of the DEP classifier $\phi$ is closer to the decision boundary of $\sgn \varepsilon_{\vetm}$ than that of $\sgn \delta_{\vetw}$. We address a good choice of the parameter $\beta$ of a DEP classifier in the following subsection. In the following subsection we also review the elegant convex-concave procedure proposed recently by Charisopoulus and Maragos to train the morphological perceptrons $\varepsilon_{\vetm}$ and $\delta_{\vetw}$ \cite{charisopoulos17}.

\subsection{Convex-Concave Procedure for Training Morphological Perceptron}

In analogy to the soft-margin support vector classifier, the weights of a morphological perceptron can be determined by solving a convex-concave optimization problem \cite{charisopoulos17,shen16}. Precisely, consider a training set $\mathcal{T} = \{(\vetx_i,d_i):i=1,\ldots,m\}$, where $\vetx_i \in \R^n$ is a training pattern and $d_i \in \{-1,+1\}$ is its binary class label for $i=1,\ldots,m$. To simplify the exposition, let $\mathcal{N}$ and $\mathcal{P}$ denote respectively the sets of negative and positive training patterns, that is, 
\begin{equation}
    \mathcal{N} = \{\vetx_i:(\vetx_i,d_i) \in \mathcal{T}, d_i = -1\} \quad
    \mbox{and} \quad 
    \mathcal{P} = \{\vetx_i:(\vetx_i,d_i) \in \mathcal{T}, d_i = +1\}.
\end{equation}
Also, let $\psi_{\boldsymbol{u}}$ be the decision function of either an erosion-based or a dilation-based morphological perceptron. In other words, let $\psi_{\boldsymbol{u}} = \varepsilon_{\vetm}$ with $\boldsymbol{u} = \vetm$ or $\psi_{\boldsymbol{u}} = \delta_{\vetw}$ with $\boldsymbol{u}=\vetw$. 
The vector $\boldsymbol{u} \in \R^n$ of either $\psi_{\boldsymbol{u}} = \varepsilon_{\vetm}$ or $\psi_{\boldsymbol{u}} = \delta_{\vetw}$ is defined as the solution of the following convex-concave optimization problem\footnote{Different from the procedured proposed by Charisopoulos and Maragos \cite{charisopoulos17}, the convex-concave optimization problem proposed in this paper includes the regularization term $C \|\boldsymbol{u}-\boldsymbol{r}\|_1$ in the objective function.}: 
\begin{align}
  \label{eq:objective}  \mathop{\mbox{minimize}}_{\boldsymbol{u},\boldsymbol{\xi}} \quad & J(\boldsymbol{u},\boldsymbol{\xi}) = 
  \frac{1}{|\mathcal{N}|}\sum_{i=1}^{|\mathcal{N}|} \nu_i^- \max\{0,\xi_i^-\} + \frac{1}{|\mathcal{P}|}\sum_{i=1}^{|\mathcal{P}|} \nu_i^+ \max\{0,\xi_i^+\} + C \|\boldsymbol{u}-\boldsymbol{r}\|_1
  ,\\
\label{eq:ineq-negative}    \mbox{subject to} \quad & \psi_{\boldsymbol{u}}(\vetx_i) \leq \xi_i^-, \quad \forall x_i \in \mathcal{N},\\
\label{eq:ineq-positive}    & \psi_{\boldsymbol{u}}(\vetx_i) \geq -\xi_i^+, \quad \forall \vetx_i \in \mathcal{P},
\end{align}
where $C$ is a regularization parameter, $\boldsymbol{r}$ is a reference value for $\boldsymbol{u}$, $\xi_i^-$ and $\xi_i^+$ are slack variable and $\nu_i^-\geq 0$ and $ \nu_i^+\geq 0$ are their penalty weights. As usual, $|\mathcal{N}|$ and $|\mathcal{P}|$ denote the cardinality of $\mathcal{N}$ and $\mathcal{P}$, respectively. 

The slack variables $\xi_i^-$ and $\xi_i^+$ measure the classification error of negative and positive training patterns weighted by $\nu_i^-$ and $\nu_i^+$, respectively. Indeed, the objective function is minimized when all slack variables are non-positive, that is, $\xi_i^- \leq 0$ and $\xi_i^+ \leq 0$ for all index $i$. On the one hand, a negative training pattern $\vetx_i \in \mathcal{N}$ is miss classified if $\psi_{\boldsymbol{u}}(\vetx_i) > 0$. From \eqref{eq:ineq-negative}, however, we have $0 < \psi_{\boldsymbol{u}}(\vetx_i) \leq \xi_i^-$ and, therefore, the objective function is not minimized. On the other hand, if a positive training pattern $\vetx_i \in \mathcal{P}$ is miss classified then $0 > \psi_{\boldsymbol{u}}(\vetx_i) \geq -\xi_i$. Equivalently, $\xi_i > 0$ and, again, the objective is not minimized.

The slack variable penalty weights $\nu_i$'s have been introduced to deal with the presence of outliers. The following presents a simple weighting scheme proposed by Charisopoulus and Maragos to penalizes training patterns with greater chances of being outliers \cite{charisopoulos17}. Let $\boldsymbol{\mu}^-$ and $\boldsymbol{\mu}^+$ be the mean of the negative and positive training patterns, that is, 
\begin{equation}
    \boldsymbol{\mu}^- = \frac{1}{|\mathcal{N}|} \sum_{\vetx_i \in \mathcal{N}} \vetx_i 
    \quad \mbox{and} \quad 
    \boldsymbol{\mu}^+ = \frac{1}{|\mathcal{P}|} \sum_{\vetx_i \in \mathcal{P}} \vetx_i,
\end{equation}
Also, let $\lambda_i^-$ and $\lambda_i^+$ be the reciprocal of the distance between $\vetx_i$ and either the mean $\boldsymbol{\mu}_i^-$ or $\boldsymbol{\mu}_i^+$. In mathematical terms,  define
\begin{equation}
    \lambda_i^- = \frac{1}{\|\vetx_i - \boldsymbol{\mu}_i^-\|}, \quad \forall \vetx_i \in \mathcal{N},
\quad \mbox{and} \quad
    \lambda_i^+ = \frac{1}{\|\vetx_i - \boldsymbol{\mu}_i^+\|}, \quad \forall \vetx_i \in \mathcal{P}.
\end{equation}
Finally, the slack variable weights $\nu_i^-$ and $\nu_i^+$ are obtained by scaling $\lambda_i^-$ and $\lambda_i^+$ to the interval $(0,1]$ as follows for all indexed $i$:
\begin{equation} \label{eq:nu-weights}
     \nu_i^- = \frac{\lambda_i^-}{\max_j \{\lambda_j^-\}}
     \quad \mbox{and} \quad 
     \nu_i^+ = \frac{\lambda_i^+}{\max_j \{\lambda_j^+\}}.
\end{equation}

As to the reference, we recommend respectively $\boldsymbol{r} = -\bigvee \mathcal{N}$ and $\boldsymbol{r} = -\bigwedge \mathcal{P}$ for the synaptic weights $\vetw$ and $\vetm$. In this case, $\delta_{\vetw}$ and $\varepsilon_{\vetm}$ classifies correctly the largest possible number of negative and positive training patterns, respectively. Also, we recommend a small regularization parameter $C$ so that the objective is dominated by the classification error measured by the slack variables. \new{Although in our computational implementation we adopted $C = 10^{-2}$, we recommend to fine tune this hyper-parameter using, for example, exhaustive search or a randomized parameter optimization strategy \cite{bergstra12}.}

Finally, we propose to train a DEP classifier using a greedy algorithm. Intuitively, the greedy algorithm first finds the best erosion-based and the best dilation-based morphological perceptrons and then it seeks for their best convex combination. Formally, we first solve two independent convex-concave optimization problems formulated using \eqref{eq:objective}-\eqref{eq:ineq-positive}, one to determine the synaptic weight $\vetm$ of the erosion-based morphological perceptron $\varepsilon_{\vetm}$ and the other to compute $\vetw$ of the dilation-based morphological perceptron $\delta_{\vetw}$. Subsequently, we determine the parameter $\beta$ by minimizing the average hinge loss. In mathematical terms, $\beta$ is obtained by solving the constrained convex problem:
\begin{equation} \label{eq:hinge-loss}
    \mathop{\mbox{minimize}}_{0 \leq \beta \leq 1} \; H(\beta) = 
  \sum_{i=1}^{m} \max\left\{0,-d_i\big[\beta \delta_{\vetw}(\vetx_i) + (1-\beta) \varepsilon_{\vetm}(\vetx_i) \big]\right\}.
\end{equation}

\begin{Remark}
In our computational experiments, we solved the optimization problem \eqref{eq:objective}-\eqref{eq:ineq-positive} using \texttt{CVXOPT} python package with the \texttt{DCCP} extension for convex-concave programing \cite{shen16} and the MOSEK solver\footnote{Further information on the \texttt{MOSEK} software package can be obtained on \url{www.mosek.com}.}. The source-code of the DEP classifier, trained using convex-concave programming and compatible with the \texttt{scikit-learn API}, is available at \url{https://github.com/mevalle/r-DEP-Classifier}.
\end{Remark}

\begin{figure}
    \centering
    \begin{tabular}{cc}
    a) Ripley's test set & b) Double-moon test set \\
    \includegraphics[width = 0.48\columnwidth]{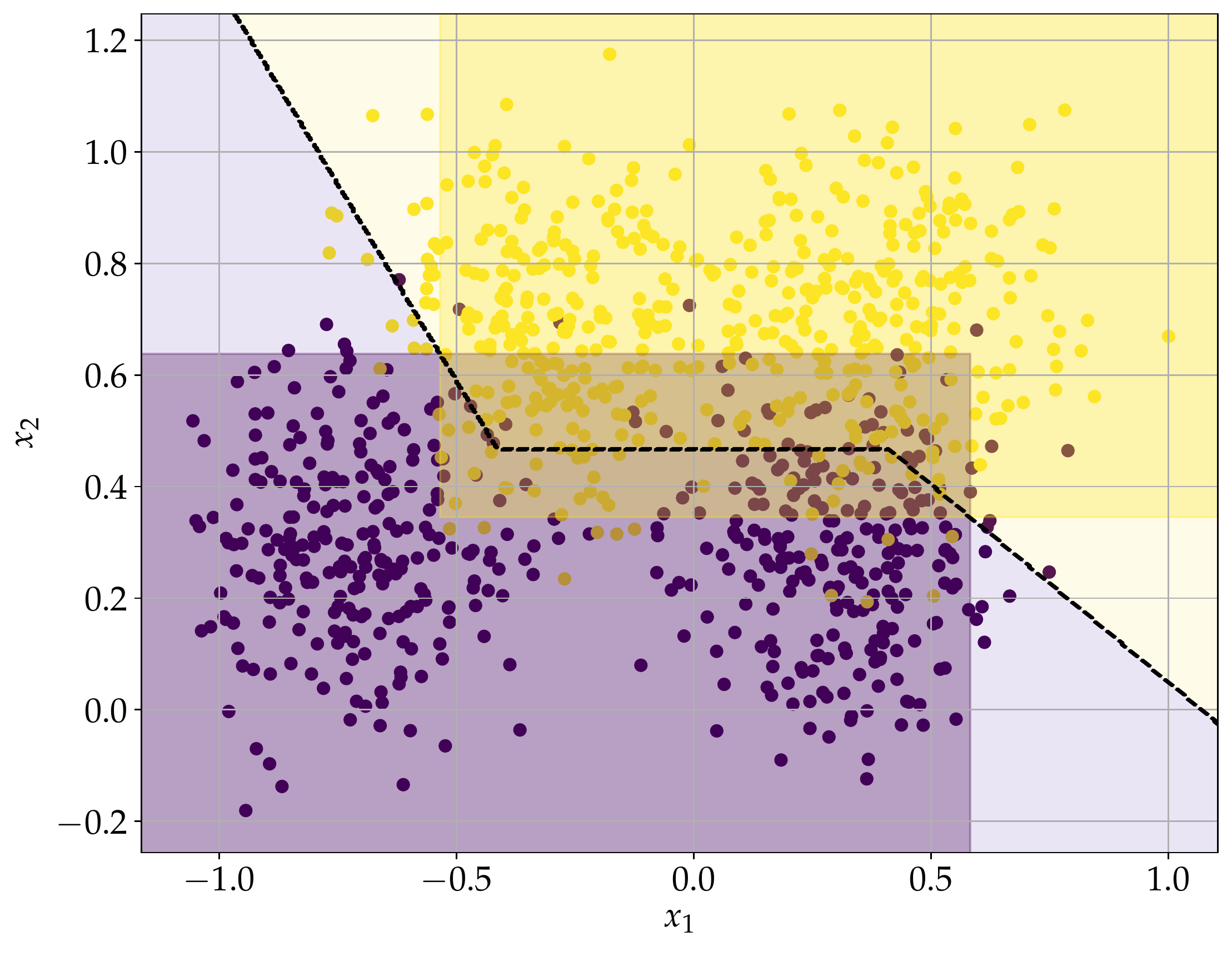} & 
    \includegraphics[width =0.48\columnwidth]{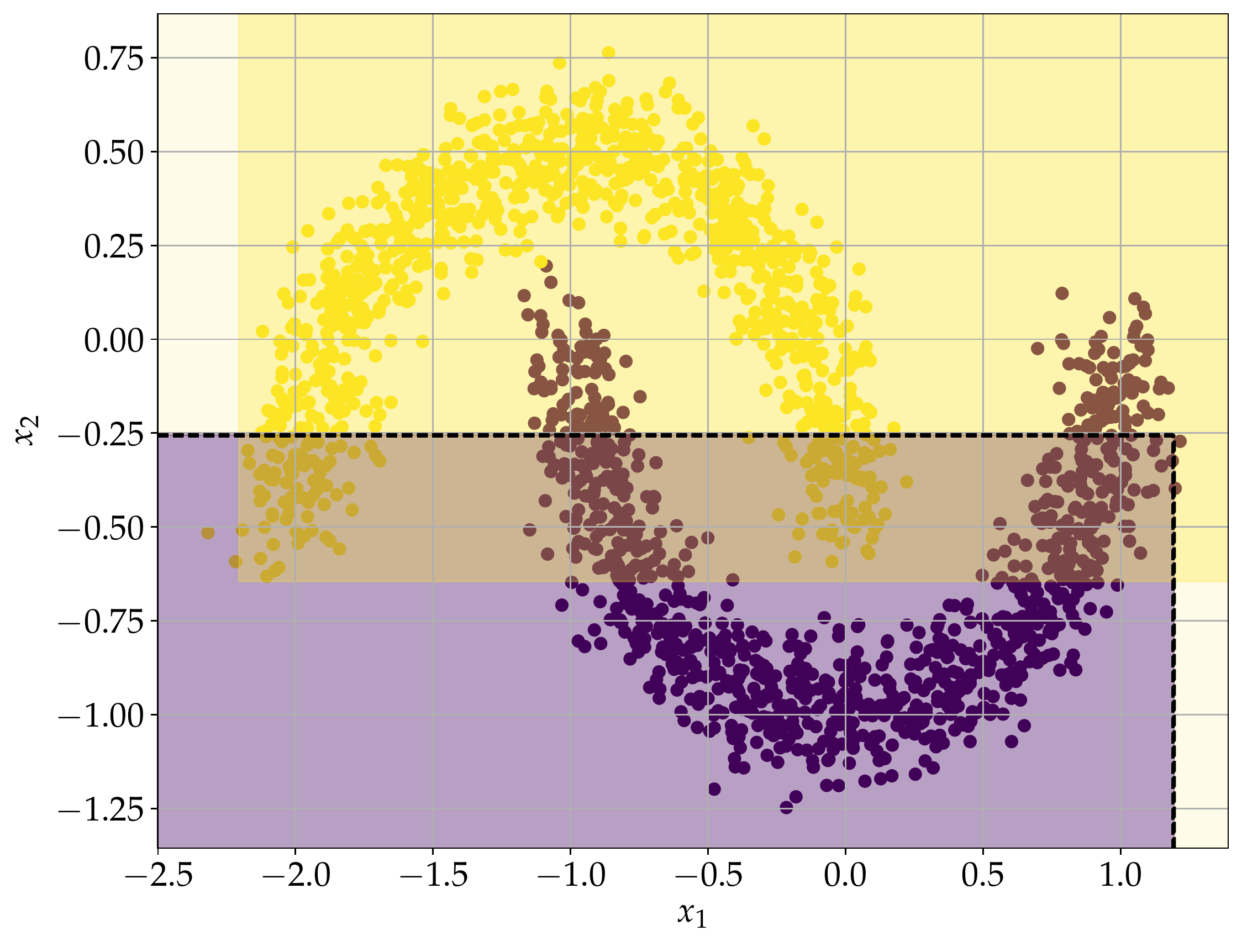}
    \end{tabular}
    \caption{Performance of the DEP classifier on Ripley's and double-moon datasets described on Examples \ref{ex:DEP-Ripley} and \ref{ex:DEP-Double-Moon}. Scatter plot of test data and the decision boundary of DEP classifier. The purple and the yellow regions corresponds respectively to the sets $D(\vetw)$ and $E(\vetm)$. }
    \label{fig:DEP}
\end{figure}

\begin{Example}[Ripley Dataset] \label{ex:DEP-Ripley}
To illustrate how the DEP classifier works, let us consider the synthetic two-class dataset of Ripley, which is already split into training and test sets \cite{ripley96}. We would like to recall that each class of Ripley's synthetic dataset has a known bimodal distribution and the best accuracy score is approximately 0.92. Using the training set with 250 samples, the convex-concave procedure yielded the synaptic weight vectors $\vetm = (0.53,-0.35)$ and $\vetw = (-0.57,-0.64)$ for $\varepsilon_{\vetm}$ and $\delta_{\vetw}$, respectively. Moreover, the convex optimization problem given by \eqref{eq:hinge-loss} yielded the parameter $\beta = 0.42$. The accuracy score on the training and test set was respectively $0.88$ and $0.90$. Figure \ref{fig:DEP}a) shows the scatter plot of the test set along with the decision boundary of the DEP classifier. This figure also depicts the regions $E(\vetm)$ and $D(\vetw)$ for the erosion-based and dilation-based morphological perceptrons.
\end{Example}
\begin{Example}[Double-Moon] \label{ex:DEP-Double-Moon}
In analogy to Haykin's dooble-moon classification problem, this problem consists of two interleaving half circles which resembles a pair of ``moons'' facing each other \cite{haykin09}. Using the command \texttt{make\_moons} from python's \texttt{scikit-learn API}, we generated training and test sets with $1,000$ and $2,000$ pairs of data, respectively. Furthermore, we corrupted both training and test data with Gaussian noise with standard variation $\sigma=0.1$. 
Figure \ref{fig:DEP}b) shows the test data together with the decision boundary of the DEP classifier and the regions $D(\vetw)$ and $E(\vetm)$ given respectively by \eqref{eq:region-D} and \eqref{eq:region-E}. In this example, the convex-concave procedure yielded the synaptic weight vectors $\vetm = (2.21,0.65)$ and $\vetw = (-1.20,0.25)$. Also, the convex optimization problem \eqref{eq:hinge-loss} yielded $\beta = 1$, which means that the DEP classifier $\sgn \tau$ coincides with the dilation-based morphological perceptron $\sgn \delta_{\vetw}$. The DEP classifier yielded an accuracy score of $0.84$ and $0.83$ for training and test sets, respectively. 
\end{Example}

Despite the DEP classifier yielded satisfactory accuracy scores is test sets from the two previous examples, this classifier has a serious drawback: As a lattice-based classifier, the DEP classifier presupposes a partial ordering on the feature space as well as on the set of classes. From \eqref{eq:morph-perceptron}, the component-wise ordering given by \eqref{eq:marginal} is adopted in the feature space while the usual total ordering of real-numbers is used to rank the class labels. Most importantly, the DEP classifier $\phi:\R^n \to \{-1,+1\}$ defined by the composition \eqref{eq:DEP-classifier} is an increasing operator because both $\sgn$ and $\tau$ are increasing operators\footnote{Note that $\tau$ is increasing because it is the convex combination of increasing operators $\varepsilon_{\vetm}$ and $\delta_{\vetw}$.}. As a consequence, the patterns from the positive class must be in general greater than the patterns from the negative class. In many practical situations, however, the component-wise ordering of the feature space is not in agreement with the natural ordering of the class labels. For example, if we invert the class labels on the synthetic dataset of Ripley, the accuracy score of the DEP classifier decreases to $0.33$ and $0.31$ for the training and test data, respectively. Similarly, the accuracy score of the DEP classifier decreases respectively to $0.66$ and $0.65$ on training and test set if we invert the class labels in the double-moon classification problem. Fortunately, we can circumvent this drawback through the use of dendrite computations \cite{ritter03}, morphological competitive units \cite{sussner11ins}, or hybrid morphological/linear neural networks \cite{hernandez19,sussner20nn}. Alternatively, we can avoid the inconsistency between the partial orderings of the feature and class spaces by making use of multi-valued mathematical morphology.

\section{Reduced Dilation-Erosion Perceptron} \label{sec:r-DEP}

As pointed out in the previous section, the DEP classifier is an increasing operator $\phi:\R^n \to \{-1,+1\}$, where the feature space $\R^n$ is equipped with the component-wise ordering given by \eqref{eq:marginal} while the set of classes $\{-1,+1\}$ inherits the natural ordering of real-numbers. In many practical situations, however, the component-wise ordering is not appropriate for the feature space. Motivated by the developments on multi-valued MM, we propose to circumvent this drawback using reduced orderings. Precisely, we introduce the so-called reduced dilation-erosion perceptron (r-DEP) which is a reduced morphological operator derived from \eqref{eq:DEP-classifier}.

Formally, let us assume the feature space is a vector-valued nonempty set $\mathbb{V}$ and let $\mathcal{C} = \{c_1,c_2\}$ be the set of classes. In practice, the feature space $\mathbb{V}$ is usually a subset of $\mathbb{R}^n$, but we may consider more abstract feature sets.  Also, let $\mathbb{L}=\bar{\R}^r$ and $\mathbb{M} = \{-1,+1\}$ be complete lattices with the component-wise ordering and the natural ordering of real numbers, respectively. 
Consider the DEP classifier $\phi: \mathbb{L} \to \mathbb{M}$ defined by \eqref{eq:DEP-classifier} for some $\vetw,\vetm \in \R^r$ and $0 \leq \beta \leq 1$. Given a one-to-one mapping $\sigma:\mathcal{C} \to \{-1,+1\}$ and a surjective mapping $\rho:\mathbb{V} \to \mathbb{L}$, from Definition \ref{def:r-MM}, the mapping $\phi^r:\mathbb{V} \to \mathcal{C}$ given by 
\begin{equation} \label{eq:r-DEP-classifier} 
    \phi^r(\vetx) = \sigma^{-1} \Big( \phi \big(\rho(\vetx) \big) \Big), \quad \forall \vetx \in \mathbb{V},
\end{equation}
is an r-increasing morphological operator because $\phi:\mathbb{L} \to \mathbb{M}$ is increasing and the identity \new{$\sigma \phi^r = \phi \rho$} holds true. Most importantly, \eqref{eq:r-DEP-classifier} defines a binary classifier $\phi^r:\mathbb{V} \to \mathcal{C}$ called {reduced dilation-erosion perceptron} (r-DEP). \new{The decision function of the r-DEP classifier is the mapping $\tau^r:\mathbb{V} \to \bar{\R}$ given by 
\begin{equation} \label{eq:r-DEP}
    \tau^r(\vetx) = \beta \delta_{\vetw}\big(\rho(\vetx)\big) + (1-\beta) \varepsilon_{\vetm}\big(\rho(\vetx)\big), \quad \forall \vetx \in \mathbb{V}.
\end{equation}}
Note that the r-DEP classifier $\phi^r$ is obtained from its decision function $\tau^r$ by means of the identity $\phi^r = \sigma^{-1} \sgn \tau^r$.

Simply put, the decision function $\tau^r$ of an r-DEP is obtained by composing the surjective mapping $\rho:\mathbb{V} \to \mathbb{L}$ and the decision function $\tau$ of a DEP, that is, $\tau^r = \tau \rho$. In other words, $\tau^r$ is obtained by applying sequentially the transformation $\rho$ and $\tau$. \new{Thus, given a training set $\mathcal{T} = \{(\vetx_i,d_i):i=1,\ldots,m\} \subset \mathbb{V} \times \mathcal{C}$, we simply train a DEP classifier using the transformed training data 
\begin{equation} \label{eq:training-r} 
\mathcal{T}^r = \{(\rho(\vetx_i),\sigma(d_i)):i=1,\ldots,m\} \subseteq \R^r \times \{-1,+1\}.
\end{equation}}
Then, the classification of an unknown pattern $\vetx \in \mathbb{V}$ is achieved by computing $\phi^r(\vetx) = \sigma^{-1}\sgn \tau \rho(\vetx)$. 

The major challenge for the design of a successful r-DEP classifier is how to determine the surjective mapping $\rho:\mathbb{V} \to \R^r$. Intuitively, the mapping $\rho$ performs a kind of dimensionality reduction which takes into account the lattice structure of patterns and labels. In this paper, we propose to determine $\rho:\mathbb{V} \to \R^d$ in a supervised manner. Specifically, based on the successful supervised reduced orderings proposed by Velasco-Forero and Angulo \cite{velasco-forero11a}, we define $\rho:\mathbb{V} \to \mathbb{L}$ using the decision function of support vector classifiers. 

Formally, \new{consider a training set $\mathcal{T} = \{(\vetx_i,d_i):i=1,\ldots,m\} \subset \mathbb{V} \times  \mathcal{C}$, where $\mathbb{V} \subset \R^n$ and $\mathcal{C}=\{c_1,c_2\}$. The mapping $\sigma:\mathcal{C} \to \{-1,+1\}$ is obtained by setting $\sigma(c_1)=-1$ and $\sigma(c_2)=+1$.} The mapping $\rho:\R^n \to \R^r$ is defined in a component-wise manner by means of the equation
$\rho(\vetx) = (\rho_1(\vetx),\rho_2(\vetx),\ldots,\rho_r(\vetx))$, where 
$\rho_1,\rho_2,\ldots,\rho_r:\R^n \to \R$ are the decision functions of distinct support vector classifiers. Recall that the decision function of a support vector classifier is given by \eqref{eq:r-SVM}.  Moreover, the distinct support vector classifiers can be determined using either one of the following approaches referred to as ensemble and bagging:
\begin{itemize}
    \item \textbf{Ensemble}: The support vector classifiers are determined using the whole training set $\mathcal{T}$ but they have different kernels.   
    \item \textbf{Bagging}: The support vector classifiers have the same kernel and parameters but they are trained using different samples of the training set $\mathcal{T}$.
\end{itemize}
The following examples, based on Ripley's and double-moon datasets, illustrate the transformation provided by these two approaches. The following examples also address the performance of the r-DEP classifier.

\begin{Example}[Ripley's Dataset] \label{ex:r-DEP-Ripley}
Consider the synthetic dataset of Ripley \cite{ripley96}.  Using the Gaussian radial basis function (RBF SVC) and the linear SVC (Linear SVC), both with the default parameters of python's \texttt{scikit-learn API}, we determined the reduced mapping $\rho$ from the training data. Figure \ref{fig:r-DEP-Ripley}a) shows the scatter plot of the transformed training set $\mathcal{T}^r$ given by \eqref{eq:training-r}. Figure \ref{fig:r-DEP-Ripley}a) also shows the regions $D(-0.59,-1.28)$ and $E(1.00,0.57)$ and the decision boundary (black-dashed-line) of the DEP classifier on the transformed space. In this example, the convex-concave optimization problem given by \eqref{eq:objective}-\eqref{eq:ineq-positive} and the minimization of the hinge loss \eqref{eq:hinge-loss} yielded $\vetm = (1.00,0.57)$, $\vetw=(-0.59,-1.28)$, and $\beta = 0.54$. Figure \ref{fig:r-DEP-Ripley}b) shows the decision boundary of the ensemble r-DEP classifier (black) on the original space together with the scatter plot of the original test set. For comparison purposes, \ref{fig:r-DEP-Ripley}b) also shows the decision boundary of the RBF-SVC (blue), linear SVC (green), and the hard-voting classifier (red) obtained using the RBF and linear SVCs. Table \ref{tab:r-DEP-Ripley} contains the accuracy score \new{(between 0 and 1)} of each of the classifiers on both training and test sets. Note that the greatest accuracy scores on the test set have been achieved by the r-DEP and the RBF-SVC classifiers. In particular, the r-DEP classifier outperformed the hard-voting ensemble classifier in this example. 
\begin{figure}[t]
    \centering
    \begin{tabular}{cc}
    a) Ensemble: Ripley's transformed training set & 
    b) Ensemble: Ripley's original test set \\
    \includegraphics[width = 0.48\columnwidth]{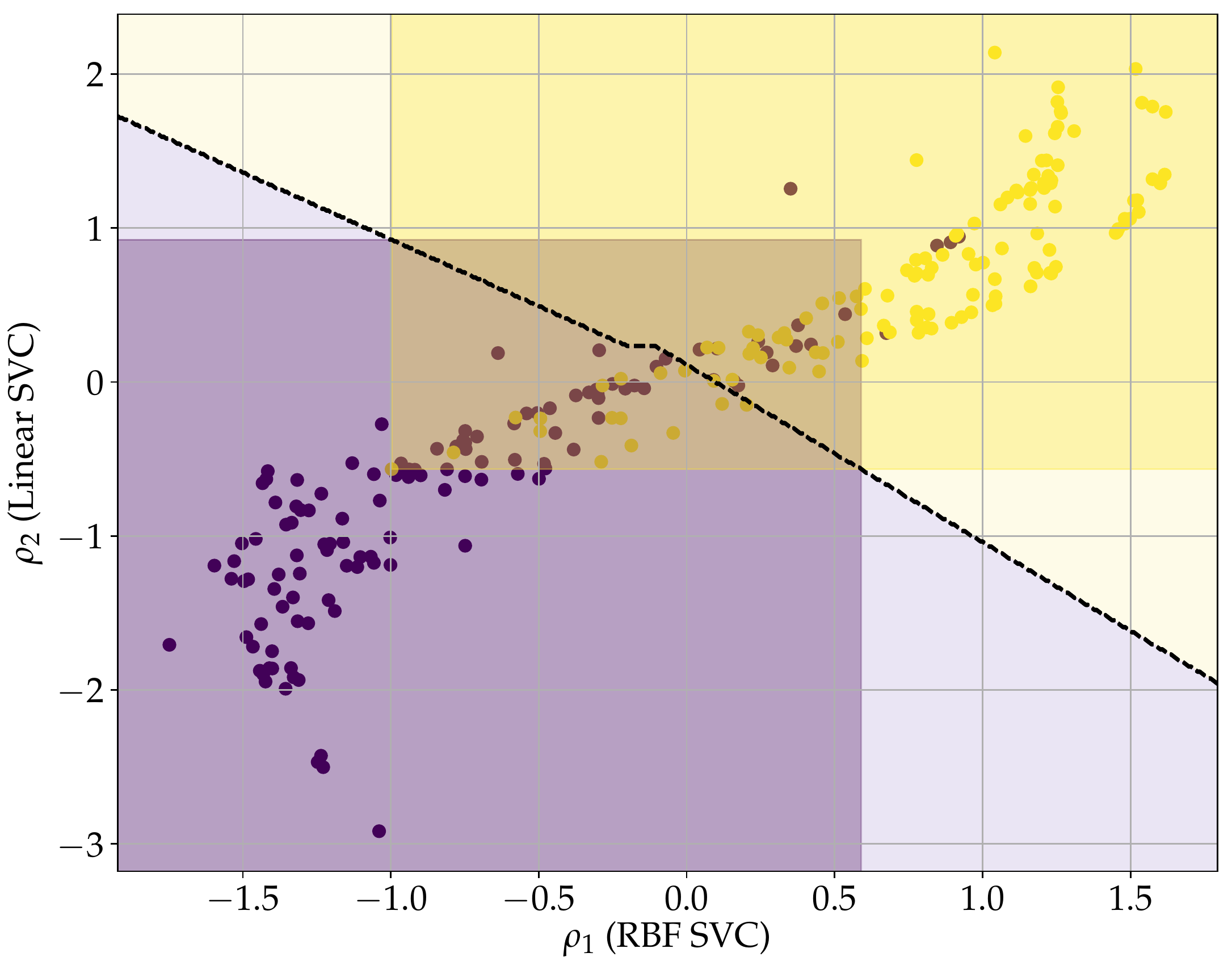} &
    \includegraphics[width = 0.48\columnwidth]{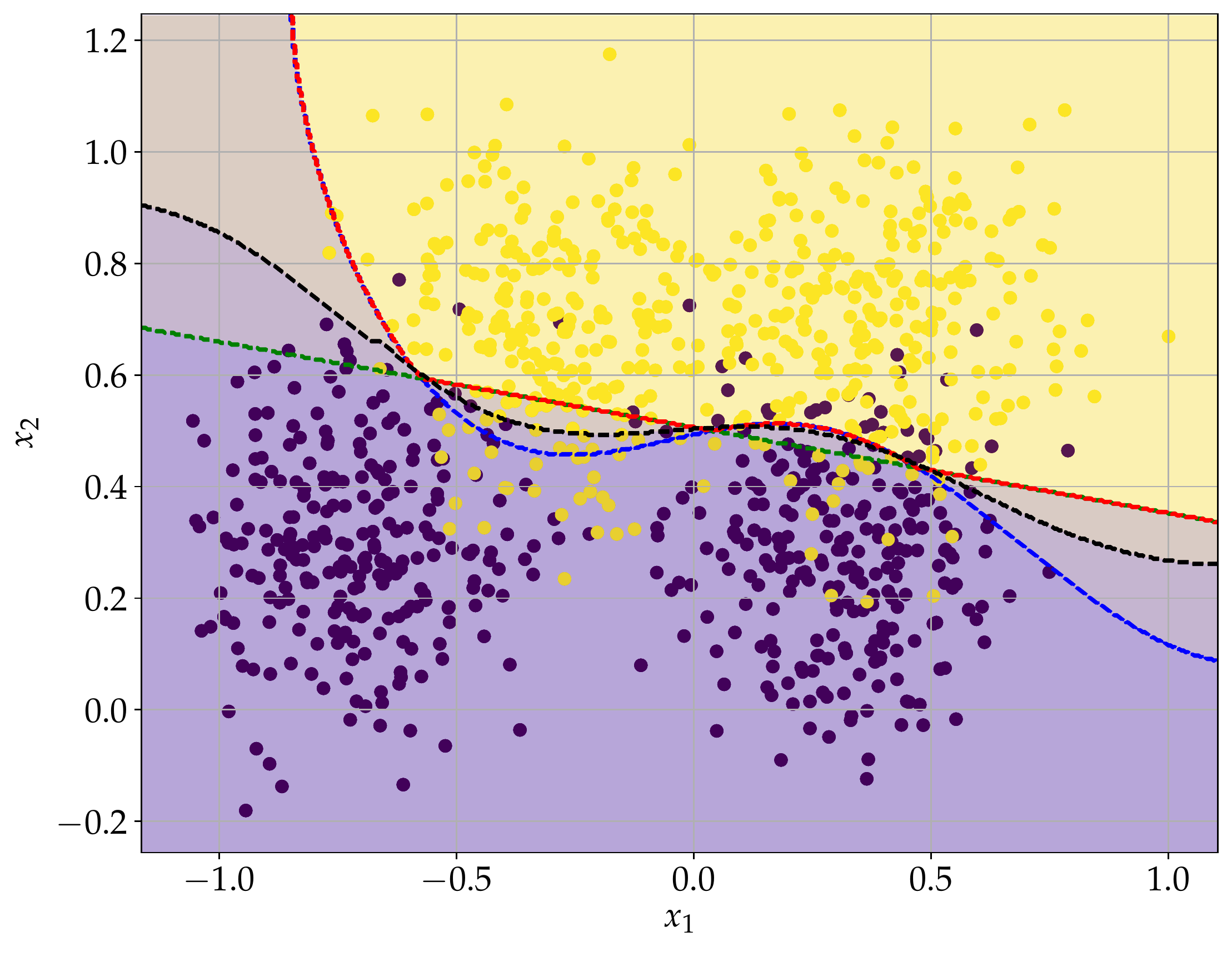} \\
    c) Bagging: Ripley's transformed training set  & 
    b) Bagging: Ripley's original test set  \\
    \includegraphics[width = 0.48\columnwidth]{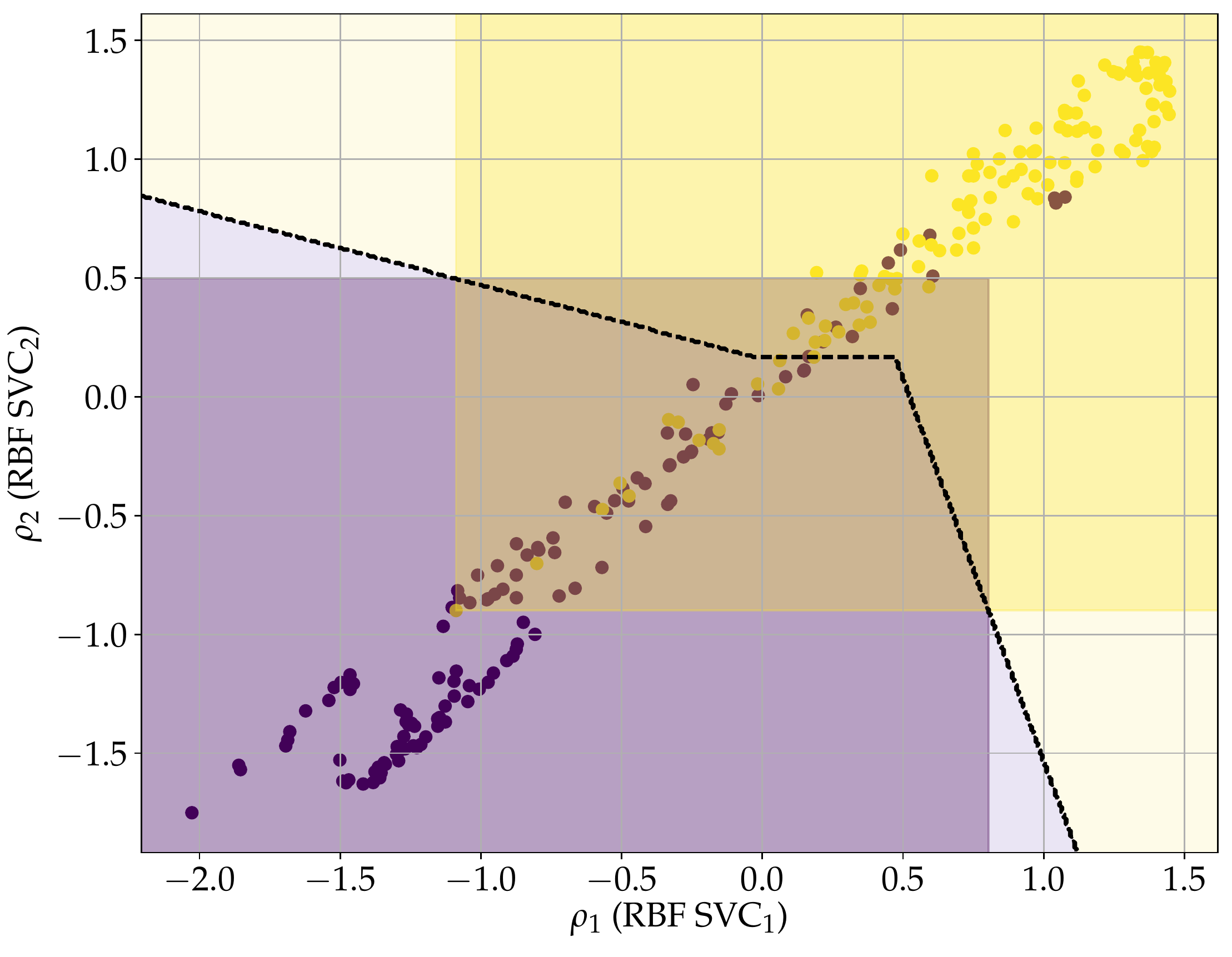} &
    \includegraphics[width = 0.48\columnwidth]{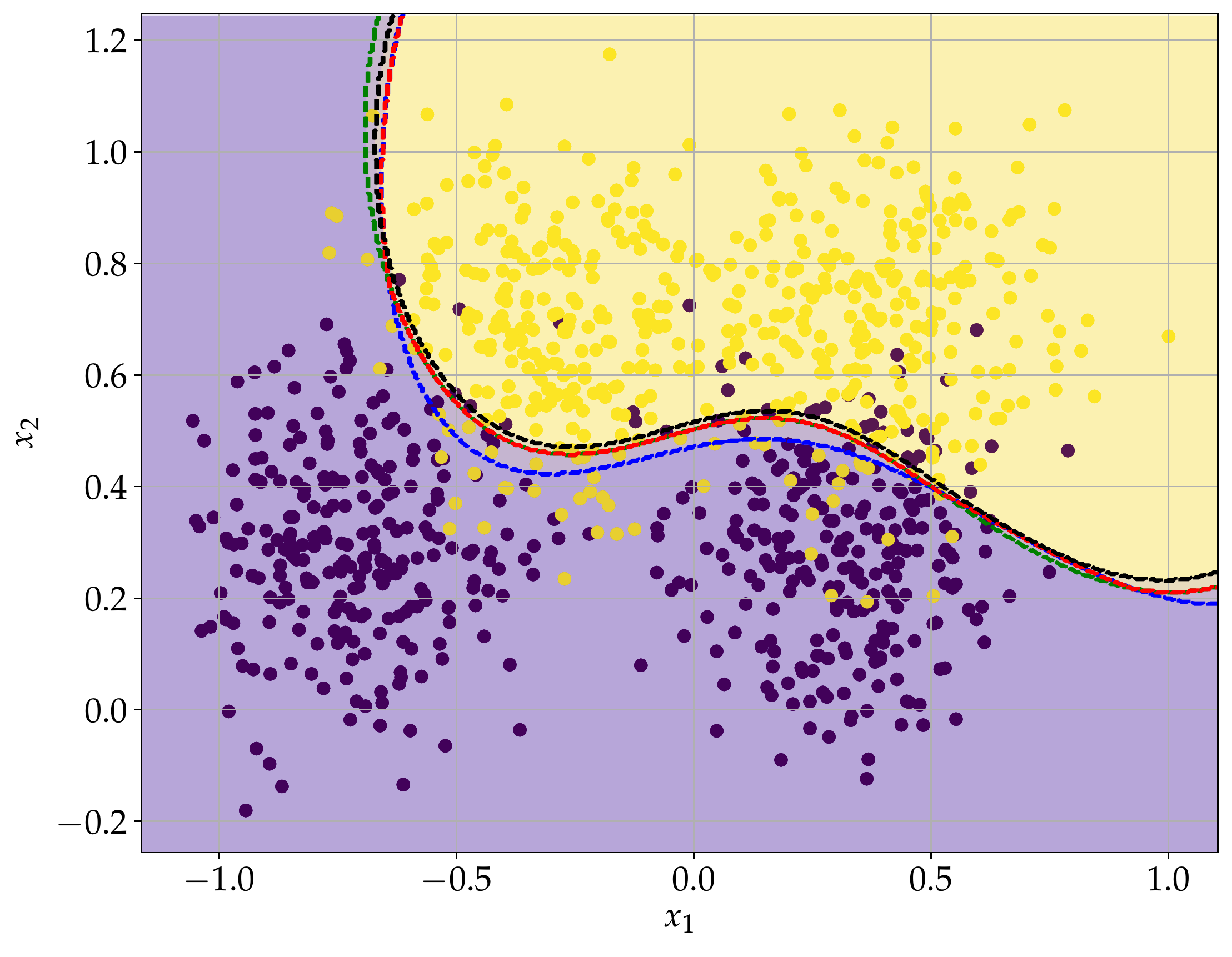}
    \end{tabular}
    \caption{Performance of ensemble and bagging r-DEP classifiers on Ripley's dataset  (see Example \ref{ex:r-DEP-Ripley}). a) and c) show the scatter plot of transformed training data, the regions $D(\vetw)$ and $E(\vetm)$, and the decision boundary of the DEP classifier. b) and d) depict the scatter plot of the original test data and the decision boundary of r-DEP (black) and other binary classifiers.}
    \label{fig:r-DEP-Ripley}
\end{figure}
\begin{table}[H]
\caption{Accuracy score of the classifiers considered in Example \ref{ex:r-DEP-Ripley} on Ripley's dataset.} \label{tab:r-DEP-Ripley}
\centering
\begin{tabular}{ccc||ccc}
\toprule
\textit{Classifier} & \textbf{Training Set}	& \textbf{TestSet} & \textit{Classifier}& \textbf{Training Set}	& \textbf{TestSet} \\
\midrule
Ensemble r-DEP &  0.86 & 0.91 & Bagging r-DEP & 0.89 &  0.90 \\
RBF SVC &  0.87 &  0.91 & RBF SVC$_1$ & 0.88 &  0.91\\
Linear SVC &  0.86 & 0.89 & RBF SVC$_2$ & 0.88 &  0.90\\
Voting SVC &  0.87	& 0.89 & Bagging SVC & 0.88 &  0.90 \\
\bottomrule
\end{tabular}
\end{table}
Similarly, we determined the mapping $\rho$ using a bagging of two distinct RBF SVCs trained with different samplings of the original training set. Precisely, we used the default parameters of a bagging classifier (\texttt{BaggingClassifier}) of the \texttt{scikit-learn} but with only two esmitamtors (\texttt{n\_estimators=2}) for a visual interpretation of the transformed data. Figure \ref{fig:r-DEP-Ripley}c) shows the scatter plot of the training data along with the regions  $D(-0.8,-0.5)$ and $E(1.09,0.90)$. In this example, the optimization problem \eqref{eq:hinge-loss} yielded $\beta = 0.76$. Figure \ref{fig:r-DEP-Ripley}d) shows the scatter plot of the original data and the decision boundaries of the classifiers: bagging r-DEP (black), RBF SVC$_1$ (blue), RBF SVC$_2$ (green), and the bagging of the two RBF SVCs (red). Table \ref{tab:r-DEP-Ripley} contains the accuracy score of these four classifiers on both training and test data. Although the RBF SVC$_1$ yielded the greatest accuracy score in the test set, the bagging r-DEP produced the largest accuracy on the training set. In general, however, the four classifiers are competitive.
\end{Example}

\begin{Example}[Double-Moon] \label{ex:r-DEP-Double-Moon}
In analogy to the previous example, we also evaluated the performance of the r-DEP classifier on the double-moon problem presented in Example \ref{ex:DEP-Double-Moon}. 
Figures \ref{fig:r-DEP-Double-Moon}a) and c) show the transformed training set obtained from the mappings determined using the ensemble and bagging strategies, respectively. We considered again a Gaussian RBF and a linear SVC in the ensemble strategy and two Gaussian RBF SVCs for the bagging. Also, we adopted the default parameters of python's \texttt{scikit-learn API} except for the number of estimators in the bagging strategy which we set to two (\texttt{n\_estimators = 2}) for a visual interpretation of the transformed data. Figure \ref{fig:r-DEP-Double-Moon}b) shows the scatter plot of the original test set with the decision boundaries of the ensemble r-DEP (black), RBF SVC (blue), linear SVC (green), and the hard-voting ensemble classifier (red). Similarly, Figure \ref{fig:r-DEP-Double-Moon}d) shows the test data with the decision boundary of the bagging r-DEP (black), RBF SVC$_1$ (blue), RBF  SVC$_2$ (green), and the bagging classifier (red). Table \ref{tab:r-DEP-Double-Moon} lists the accuracy score \new{(between 0 and 1)} of all the classifiers on both training and test sets of the double-moon problem. As expected, the linear SVC yielded the worst perforamnce. The largest scores have been achieved by both ensemble and bagging r-DEP as well as the Gaussian RBF SVCs and their bagging.
\begin{figure}[t]
    \centering
    \begin{tabular}{cc}
    a) Ensemble: Double-moon transformed training set & 
    b) Ensemble: Double-moon original test set \\
    \includegraphics[width = 0.48\columnwidth]{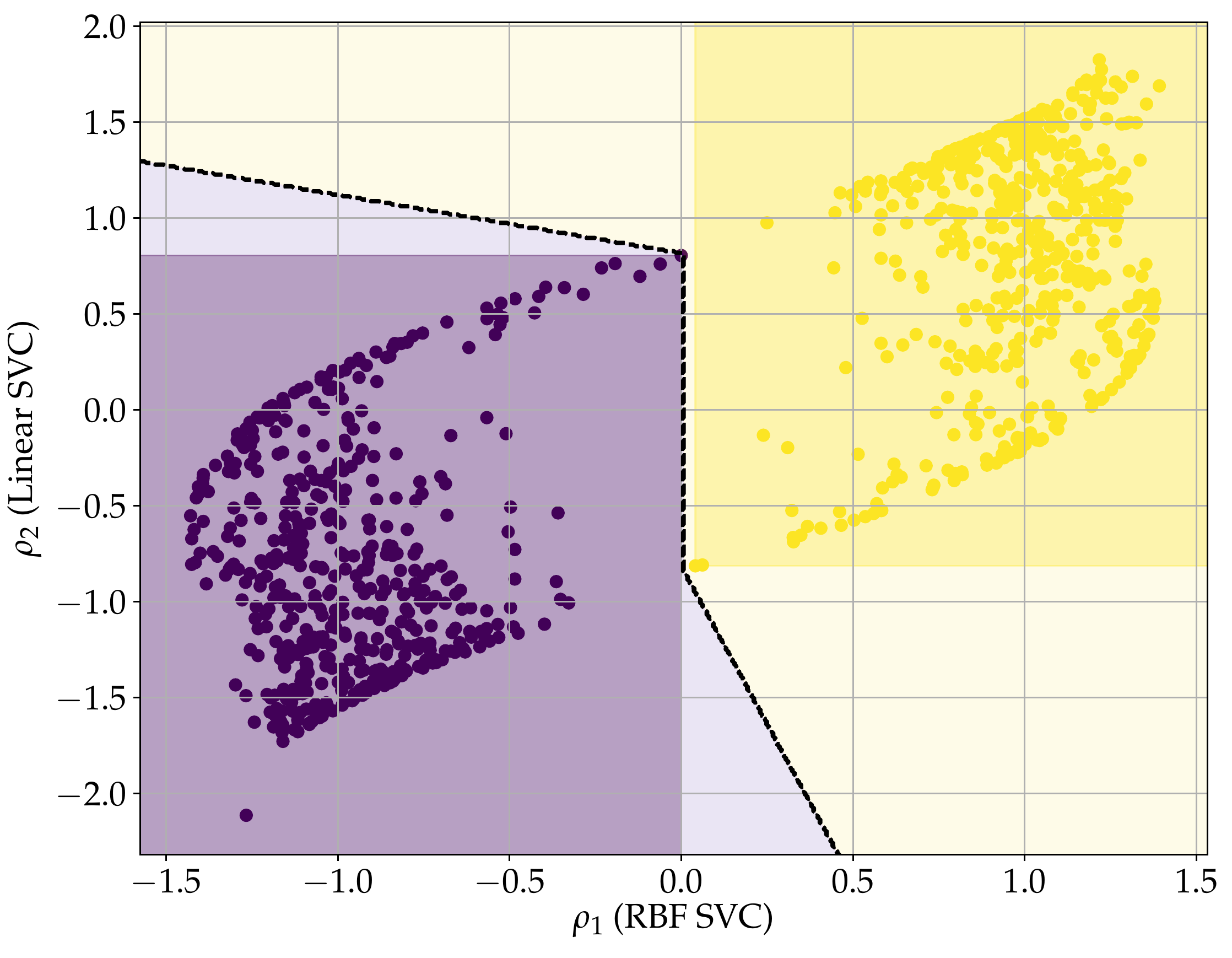} &
    \includegraphics[width = 0.48\columnwidth]{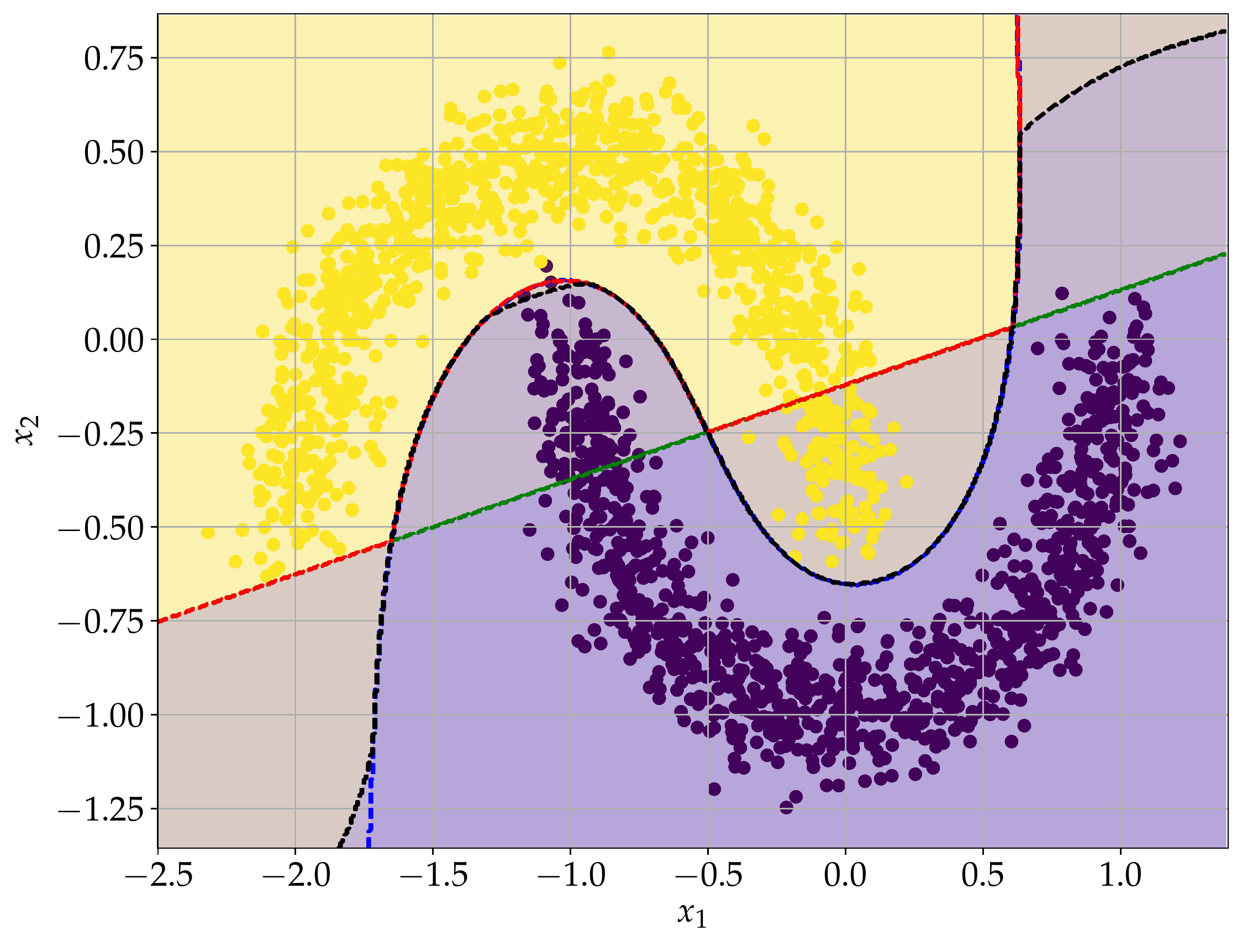} \\
    c) Bagging: Double-moon transformed training set  & 
    b) Bagging: Double-moon original test set  \\
    \includegraphics[width = 0.48\columnwidth]{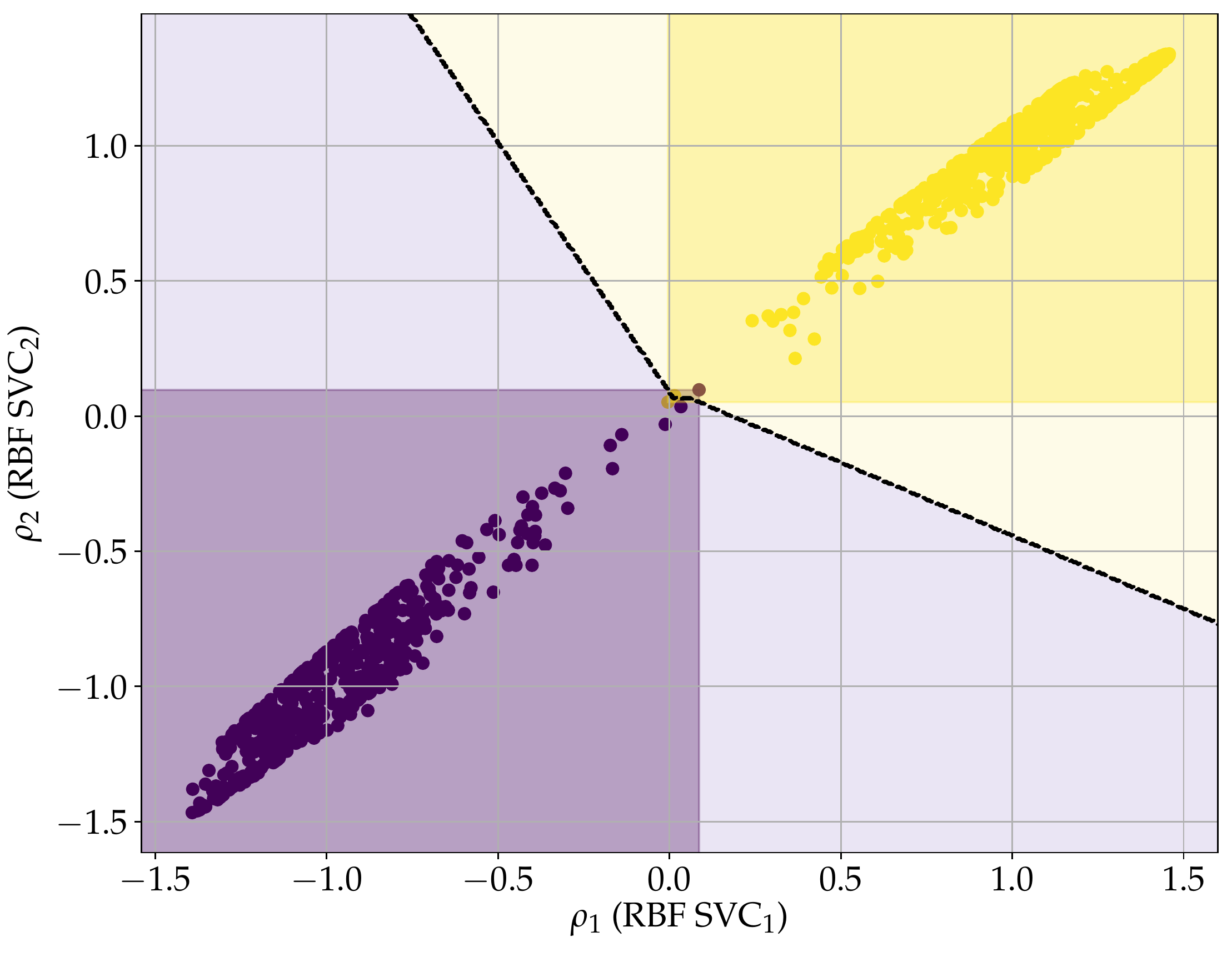} &
    \includegraphics[width = 0.48\columnwidth]{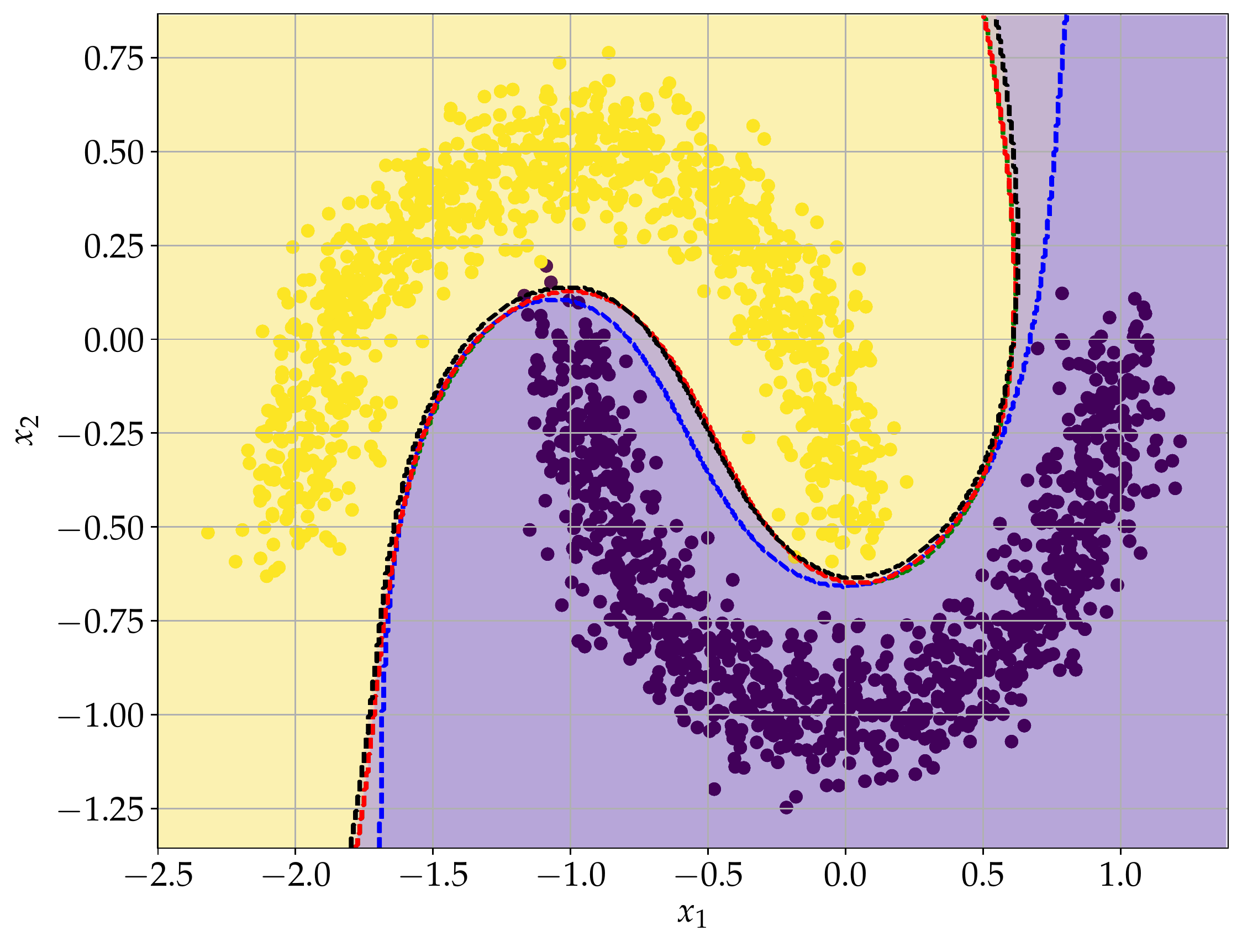}
    \end{tabular}
    \caption{Performance of the ensemble r-DEP classifier on double-moon datasets described on Example \ref{ex:r-DEP-Double-Moon}. a) and c) show the scatter plot of transformed training data, the regions $D(\vetw)$ and $E(\vetm)$, and the decision boundary of the DEP classifier. b) and d) depict the scatter plot of the original test data and the decision boundary of r-DEP (black) and other binary classifiers.}
    \label{fig:r-DEP-Double-Moon}
\end{figure}
\begin{table}[H]
\caption{Accuracy score of the classifiers considered in Example \ref{ex:r-DEP-Double-Moon} on double-moon problem.} \label{tab:r-DEP-Double-Moon}
\centering
\begin{tabular}{ccc||ccc}
\toprule
\textit{Classifier} & \textbf{Training Set}	& \textbf{TestSet} & \textit{Classifier}& \textbf{Training Set}	& \textbf{TestSet} \\
\midrule
Ensemble r-DEP &  1.00 & 1.00 & 
Bagging r-DEP & 1.00 &  1.00 \\
RBF SVC &  1.00 &  1.00 & RBF SVC$_1$ & 1.00 &  1.00\\
Linear SVC &  0.88 & 0.88 & RBF SVC$_2$ & 1.00 & 1.00\\
Voting SVC &  0.93	& 0.94 & Bagging SVC & 1.00 &  1.00 \\
\bottomrule
\end{tabular}
\end{table}
\end{Example}

We would like to point out that, in contrast to the original DEP classifier, the perforamance of the r-DEP model remains high if we change the pattern labels. In the following section we provide more conclusive computational experiments concerning the performance of r-DEP for binary classification.

\section{Computational Experiments} \label{sec:Experiments}

Let us now provide extensive computational experiments to evaluate the performance of the ensemble and bagging r-DEP classifiers. In the ensemble strategy, the mapping $\rho$ is obtained by considering a RBF SVC, a linear SVC, and a polynomial SVC. The bagging strategy consists of 10 RBF SVCs where each base estimator has been trained using a sampling of the original training set with replacement. Let us also compare the new r-DEP classifiers with the original DEP classifier, linear SVC, RBF SVC, the polynomial SVC (poly SVC) as well as an ensemble of the three SVCs and a bagging of RBF SVCs. We would like to point out that we used the default parameters of the python's \texttt{scikit-learn API} in our computational experiments \cite{scikit-learn,sklearn_api}. 

We considered a total of 30 binary classification problems available on the OpenML repository available at \url{https://www.openml.org/} \cite{OpenML}. We would like to point out that most datasets we considered are also available at the well-known UCI machine learning repository \cite{uci_repository}. We used the OpenML repository because all the datasets can be accessed by means of the command \texttt{fetch\_openml} from python's \texttt{scikit-learn} \cite{scikit-learn}. Moreover, we handled missing data using the \texttt{SimpleImputer} command, also from \texttt{scikit-learn}. Table \ref{tab:datasets} lists the 30 datasets considered. Table \ref{tab:datasets} also include the number of instances ($\#$instances), the number of features ($\#$features), the percentage of the negative and positive patterns, denoted by the pair $(\mathcal{N}\%,\mathcal{P}\%)$, and the OpenML name/version.
\begin{table}[t]
\caption{Informations on the considered datasets.} \label{tab:datasets}
\centering
\begin{tabular}{cccccc}
\toprule & \textbf{Name} & \textbf{\texttt{\#} Instances} & \textbf{\texttt{\#} Features} & $\mathbf{ (\mathcal{N}\%,\mathcal{P}\%)}$ & \textbf{OpenML Name/Version} \\
\midrule
1 & Arsene & 200 & 10000 & (44,56) & arcene/1 \\ 
2 & Australian & 690 & 14 & (56,44) & Australian/4 \\ 
3 & Banana & 5300 & 2 & (55,45) & banana/1 \\ 
4 & Banknote & 1372 & 4 & (56,44) & banknote-authentication/1 \\ 
5 & Blood Transfusion & 748 & 4 & (76,24) & blood-transfusion-service-center/1 \\ 
6 & Breast Cancer Wisconsin & 569 & 30 & (63,37) & wdbc/1 \\
7 & Chess & 3196 & 36 & (48,52) & kr-vs-kp/1 \\ 
8 & Colic & 368 & 22 & (37,63) & colic/2 \\ 
9 & Credit Approval & 690 & 15 & (44,56) & credit-approval/1 \\ 
10 & Credit-g & 1000 & 20 & (30,70) & credit-g/1 \\ 
11 & Cylinder Bands & 540 & 37 & (42,58) & cylinder-bands/2 \\ 
12 & Diabetes & 768 & 8 & (65,35) & diabetes/1 \\ 
13 & Egg-Eye-State & 14980 & 14 & (55,45) & eeg-eye-state/1 \\ 
14 & Haberman & 306 & 3 & (74,26) & haberman/1 \\ 
15 & Hill-Valley & 1212 & 100 & (50,50) & hill-valley/1 \\ 
16 & Ilpd & 583 & 10 & (71,29) & ilpd/1 \\ 
17 & Internet Advertisements & 3279 & 1558 & (14,86) & Internet-Advertisements/2 \\
18 & Ionosphere & 351 & 34 & (36,64) & ionosphere/1 \\ 
19 & MOFN-3-7-10 & 1324 & 10 & (22,78) & mofn-3-7-10/1 \\ 
20 & Monks-2 & 601 & 6 & (66,34) & monks-problems-2/1 \\ 
21 & Mushroom & 8124 & 22 & (52,48) & mushroom/1 \\ 
22 & Phoneme & 5404 & 5 & (71,29) & phoneme/1 \\ 
23 & Pishing Websites & 11055 & 30 & (44,56) & PhishingWebsites/1 \\ 
24 & Sick & 3772 & 29 & (94, 6) & sick/1 \\ 
25 & Sonar & 208 & 60 & (53,47) & sonar/1 \\ 
26 & Spambase & 4601 & 57 & (61,39) & spambase/1 \\ 
27 & Steel Plates Fault & 1941 & 33 & (65,35) & steel-plates-fault/1 \\ 
28 & Thoracic Surgery & 470 & 16 & (85,15) & thoracic$\_$surgery/1 \\ 
29 & Tic-Tac-Toe & 958 & 9 & (35,65) & tic-tac-toe/1 \\ 
30 & Titanic & 2201 & 3 & (68,32) & Titanic/2 \\   
\bottomrule
\end{tabular}
\end{table}

Note that the number of samples ranges from 200 (Arsene) to 14,980 (Egg-Eye-State) while the number of features varies from 2 (Banana) to 10,000 (Arsene). Furthermore, some datasets such as the \textit{Sick} and \textit{Toracic Surgery} are extremely unbalanced. \new{Therefore, we used the balanced accuracy score, which ranges from 0 to 1, to measure the performance of a classifier \cite{brodersen10}.} Table \ref{tab:accuracies} contain the mean and standard deviation of the balanced accuracy score obtained using a stratified $10$-fold cross-validation. The largest mean score for each dataset have been typed using boldface. 

We would like to point out that, to avoid biases, we used the same training and test partition for all the classifiers. Also, we pre-processed the data using the command \texttt{StandardScaler} from \texttt{scikit-learn}, that is, we computed the mean and the standard deviation of each feature on the training set and normalized both training and test sets using the obtained values. The \texttt{StandardScaler} transformation has also been applied on the output of the $\rho$ mapping. The source-code of the computational experiment is available at \url{https://github.com/mevalle/r-DEP-Classifier}. 

\begin{figure}[t]
    \centering
    \includegraphics[width = \columnwidth]{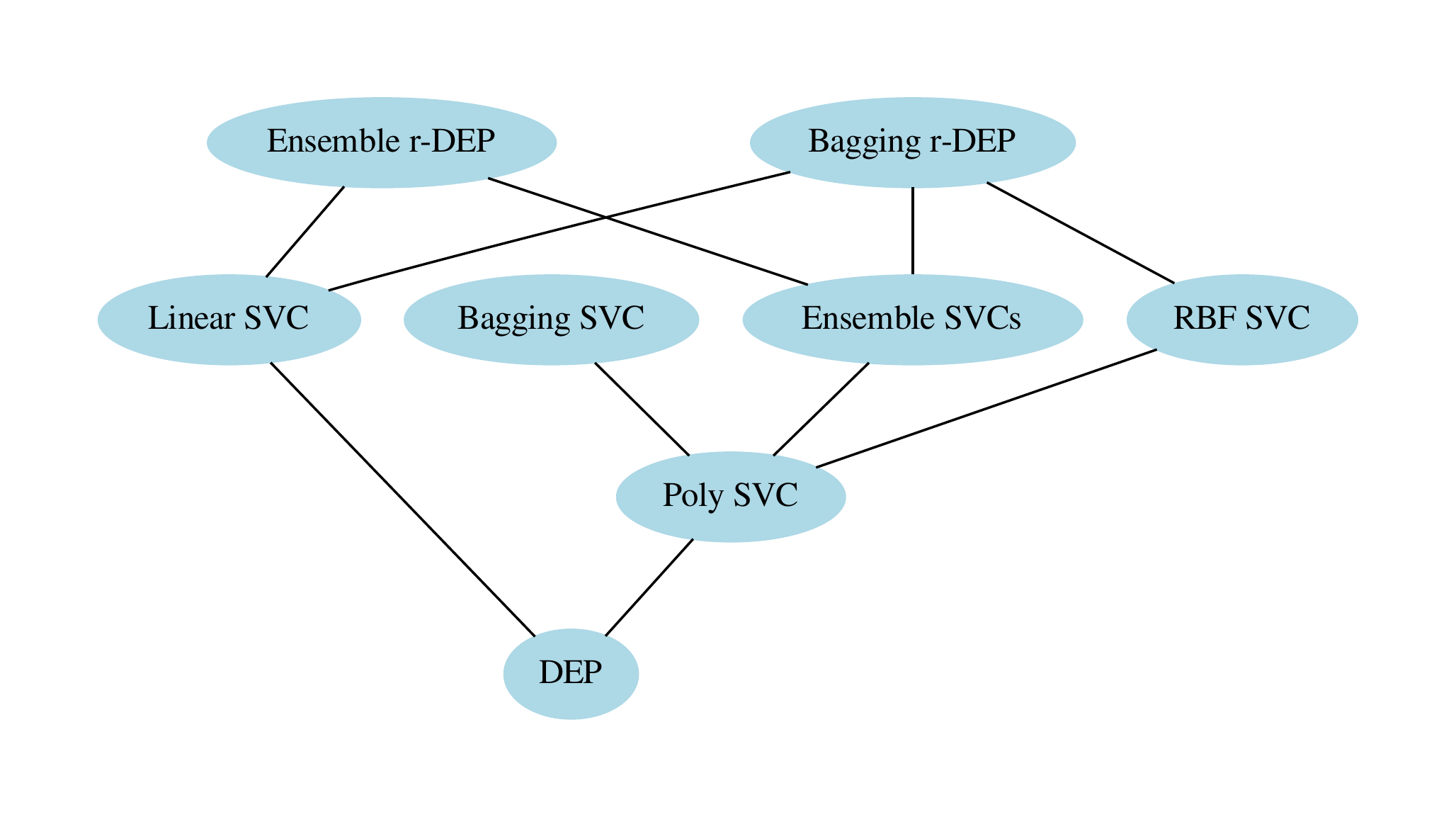}
    \caption{Hasse diagram of paired Student's t-test with confidence level at 99\%.}
    \label{fig:HoTDiagram}
\end{figure}

From Table \ref{tab:accuracies}, the largest average of the balanced accuracy scores have been achieved by the ensemble and bagging r-DEP classifiers. Using paired Student's t-test with confidence level at 99\%, we confirmed that the ensemble and bagging r-DEP, in general, performed better than the other classifiers. \new{In fact, Figure \ref{fig:HoTDiagram} shows the Hasse diagram of the outcome of paired hypothesis tests \cite{burda13,weise15}. Specifically, an edge in this diagram means that the hypothesis test discarded the null hypothesis that the classifier on the top yielded balanced accuracy score less than or equal to the classifier on the bottom. For example, Student’s t-test discarded the null hypothesis that the ensemble r-DEP classifier performs as well as or worst than the hard-voting ensemble of SVCs. In other words, the ensemble r-DEP statistically outperformed the ensemble of SVCs. Concluding, in Figure \ref{fig:HoTDiagram}, the method on the top of an edge statistically outperformed the method on the bottom. }

\begin{table}
\caption{Average and standard deviation of the balanced accuracy scores obtained from the classifiers using stratified 10-fold cross-validation.} \label{tab:accuracies}
\centering
\rotatebox{90}{
\begin{tabular}{ccccccccc}
\toprule 
 & \parbox{0.08\columnwidth}{\centering \textbf{Linear SVC}} & \parbox{0.08\columnwidth}{\centering \textbf{RBF SVC}} & \parbox{0.08\columnwidth}{\centering \textbf{Poly SVC}} & 
 \parbox{0.08\columnwidth}{\centering \textbf{Ensemble of SVCs}} & \parbox{0.08\columnwidth}{\centering \textbf{Bagging of SVC}} & \textbf{DEP} & \parbox{0.08\columnwidth}{\centering \textbf{Ensemble r-DEP}} & 
 \parbox{0.08\columnwidth}{\centering \textbf{Bagging r-DEP}} \\ 
 \midrule
Arsene & $\mathbf{0.90\pm0.08}$ & 0.77$\pm$0.05 & 0.77$\pm$0.10 & 0.85$\pm$0.07 & 0.79$\pm$0.07 & 0.53$\pm$0.05 & 0.88$\pm$0.08 & 0.77$\pm$0.08 \\ 
Australian & 0.86$\pm$0.03 & 0.86$\pm$0.04 & 0.84$\pm$0.04 & 0.85$\pm$0.05 & $\mathbf{0.86\pm0.04}$ & 0.72$\pm$0.14 & 0.86$\pm$0.04 & 0.86$\pm$0.04 \\ 
Banana & 0.50$\pm$0.00 & 0.90$\pm$0.01 & 0.61$\pm$0.02 & 0.63$\pm$0.01 & $\mathbf{0.90\pm0.01}$ & 0.48$\pm$0.02 & 0.89$\pm$0.01 & 0.90$\pm$0.01 \\ 
Banknote & 0.99$\pm$0.01 & $\mathbf{1.00\pm0.00}$ & 0.99$\pm$0.01 & 0.99$\pm$0.01 & $\mathbf{1.00\pm0.00}$ & 0.41$\pm$0.05 & $\mathbf{1.00\pm0.00}$ & $\mathbf{1.00\pm0.00}$ \\ 
Blood Transfusion & 0.50$\pm$0.00 & 0.56$\pm$0.03 & 0.53$\pm$0.03 & 0.53$\pm$0.02 & 0.55$\pm$0.03 & 0.55$\pm$0.06 & $\mathbf{0.66\pm0.06}$ & 0.55$\pm$0.06 \\ 
Breast Cancer Wisconsin & 0.97$\pm$0.02 & 0.97$\pm$0.02 & 0.87$\pm$0.03 & 0.97$\pm$0.01 & $\mathbf{0.97\pm0.02}$ & 0.88$\pm$0.06 & 0.97$\pm$0.02 & 0.97$\pm$0.02 \\ 
Chess & 0.97$\pm$0.01 & 0.99$\pm$0.01 & 0.97$\pm$0.01 & 0.99$\pm$0.01 & 0.99$\pm$0.01 & 0.52$\pm$0.03 & 0.99$\pm$0.01 & $\mathbf{0.99\pm0.01}$ \\ 
Colic & 0.81$\pm$0.07 & 0.83$\pm$0.07 & 0.70$\pm$0.05 & 0.82$\pm$0.07 & $\mathbf{0.84\pm0.08}$ & 0.56$\pm$0.10 & 0.81$\pm$0.05 & 0.82$\pm$0.07 \\ 
Credit Approval & 0.86$\pm$0.04 & 0.86$\pm$0.04 & 0.85$\pm$0.05 & $\mathbf{0.87\pm0.05}$ & 0.86$\pm$0.04 & 0.55$\pm$0.13 & 0.79$\pm$0.08 & 0.85$\pm$0.04 \\ 
Credit-g & 0.67$\pm$0.06 & 0.67$\pm$0.07 & 0.61$\pm$0.05 & 0.66$\pm$0.07 & 0.67$\pm$0.07 & 0.55$\pm$0.09 & 0.69$\pm$0.07 & $\mathbf{0.71\pm0.06}$ \\ 
Cylinder Bands & 0.66$\pm$0.07 & 0.77$\pm$0.07 & 0.66$\pm$0.09 & 0.71$\pm$0.07 & 0.75$\pm$0.08 & 0.53$\pm$0.05 & $\mathbf{0.79\pm0.06}$ & 0.76$\pm$0.09 \\ 
Diabetes & 0.73$\pm$0.07 & 0.70$\pm$0.07 & 0.66$\pm$0.05 & 0.71$\pm$0.06 & 0.71$\pm$0.06 & 0.65$\pm$0.05 & $\mathbf{0.74\pm0.05}$ & 0.72$\pm$0.04 \\ 
Egg-Eye-State & 0.58$\pm$0.02 & 0.63$\pm$0.05 & 0.53$\pm$0.01 & 0.59$\pm$0.03 & 0.63$\pm$0.05 & 0.49$\pm$0.03 & 0.59$\pm$0.03 & $\mathbf{0.66\pm0.04}$ \\ 
Haberman & 0.50$\pm$0.02 & 0.57$\pm$0.08 & 0.50$\pm$0.02 & 0.50$\pm$0.02 & 0.56$\pm$0.05 & 0.58$\pm$0.07 & $\mathbf{0.60\pm0.11}$ & 0.58$\pm$0.12 \\ 
Hill-Valley & 0.61$\pm$0.03 & 0.50$\pm$0.04 & 0.53$\pm$0.02 & 0.58$\pm$0.03 & 0.50$\pm$0.04 & 0.52$\pm$0.05 & $\mathbf{0.69\pm0.11}$ & 0.52$\pm$0.04 \\ 
Ilpd & 0.50$\pm$0.00 & 0.51$\pm$0.02 & 0.50$\pm$0.02 & 0.50$\pm$0.01 & 0.50$\pm$0.02 & 0.46$\pm$0.08 & $\mathbf{0.63\pm0.09}$ & 0.61$\pm$0.09 \\
Internet Advertisements & 0.91$\pm$0.04 & 0.89$\pm$0.02 & 0.81$\pm$0.04 & 0.88$\pm$0.03 & 0.89$\pm$0.03 & 0.77$\pm$0.08 & $\mathbf{0.93\pm0.03}$ & 0.89$\pm$0.03 \\ 
Ionosphere & 0.86$\pm$0.06 & 0.93$\pm$0.04 & 0.67$\pm$0.07 & 0.86$\pm$0.07 & 0.94$\pm$0.04 & 0.90$\pm$0.07 & 0.88$\pm$0.05 & $\mathbf{0.94\pm0.02}$ \\ 
MOFN-3-7-10 & $\mathbf{1.00\pm0.00}$ & $\mathbf{1.00\pm0.00}$ & $\mathbf{1.00\pm0.00}$ & $\mathbf{1.00\pm0.00}$ & $\mathbf{1.00\pm0.00}$ & 0.57$\pm$0.07 & $\mathbf{1.00\pm0.00}$ & $\mathbf{1.00\pm0.00}$ \\ 
Monks-2 & 0.50$\pm$0.00 & 0.65$\pm$0.06 & 0.49$\pm$0.02 & 0.50$\pm$0.01 & 0.65$\pm$0.06 & 0.49$\pm$0.03 & 0.70$\pm$0.06 & $\mathbf{0.79\pm0.04}$ \\ 
Mushroom & 0.98$\pm$0.01 & $\mathbf{1.00\pm0.00}$ & $\mathbf{1.00\pm0.00}$ & $\mathbf{1.00\pm0.00}$ & $\mathbf{1.00\pm0.00}$ & 0.71$\pm$0.11 & $\mathbf{1.00\pm0.00}$ & $\mathbf{1.00\pm0.00}$ \\ 
Phoneme & 0.72$\pm$0.02 & 0.81$\pm$0.02 & 0.65$\pm$0.02 & 0.74$\pm$0.02 & 0.81$\pm$0.02 & 0.74$\pm$0.02 & 0.82$\pm$0.03 & $\mathbf{0.83\pm0.03}$ \\ 
Pishing Websites & 0.90$\pm$0.01 & 0.95$\pm$0.01 & 0.94$\pm$0.01 & 0.94$\pm$0.01 & 0.95$\pm$0.01 & 0.59$\pm$0.10 & 0.95$\pm$0.01 & $\mathbf{0.95\pm0.01}$ \\ 
Sick & 0.81$\pm$0.06 & 0.75$\pm$0.02 & 0.61$\pm$0.04 & 0.75$\pm$0.05 & 0.75$\pm$0.03 & 0.52$\pm$0.04 & $\mathbf{0.89\pm0.03}$ & 0.88$\pm$0.05 \\ 
Sonar & 0.75$\pm$0.09 & 0.83$\pm$0.07 & 0.82$\pm$0.06 & 0.82$\pm$0.07 & 0.84$\pm$0.08 & 0.49$\pm$0.09 & 0.85$\pm$0.08 & $\mathbf{0.87\pm0.07}$ \\ 
Spambase & 0.92$\pm$0.01 & 0.93$\pm$0.01 & 0.73$\pm$0.02 & 0.93$\pm$0.01 & 0.93$\pm$0.01 & 0.51$\pm$0.06 & 0.93$\pm$0.01 & $\mathbf{0.93\pm0.01}$ \\ 
Steel Plates Fault & $\mathbf{1.00\pm0.00}$ & 1.00$\pm$0.01 & 1.00$\pm$0.00 & $\mathbf{1.00\pm0.00}$ & 1.00$\pm$0.01 & 0.51$\pm$0.02 & $\mathbf{1.00\pm0.00}$ & 1.00$\pm$0.00 \\ 
Thoracic Surgery & 0.50$\pm$0.00 & 0.50$\pm$0.00 & 0.49$\pm$0.02 & 0.50$\pm$0.00 & 0.50$\pm$0.00 & 0.51$\pm$0.10 & $\mathbf{0.56\pm0.12}$ & 0.50$\pm$0.10 \\ 
Tic-Tac-Toe & 0.50$\pm$0.00 & 0.86$\pm$0.05 & 0.74$\pm$0.06 & 0.74$\pm$0.06 & 0.87$\pm$0.05 & 0.61$\pm$0.05 & $\mathbf{0.89\pm0.04}$ & 0.89$\pm$0.03 \\ 
Titanic & 0.70$\pm$0.03 & 0.68$\pm$0.02 & $\mathbf{0.71\pm0.03}$ & 0.71$\pm$0.03 & 0.69$\pm$0.02 & 0.43$\pm$0.09 & 0.60$\pm$0.06 & 0.70$\pm$0.03 \\ \midrule
\textbf{Average} & 0.76$\pm$0.18 & 0.80$\pm$0.16 & 0.73$\pm$0.17 & 0.77$\pm$0.17 & 0.80$\pm$0.16 & 0.58$\pm$0.12 & $\mathbf{0.82\pm0.14}$ & 0.81$\pm$0.15 \\ 
\bottomrule
\end{tabular} }
\end{table}

\begin{figure}[t]
    \centering
    \includegraphics[width = 0.95\columnwidth]{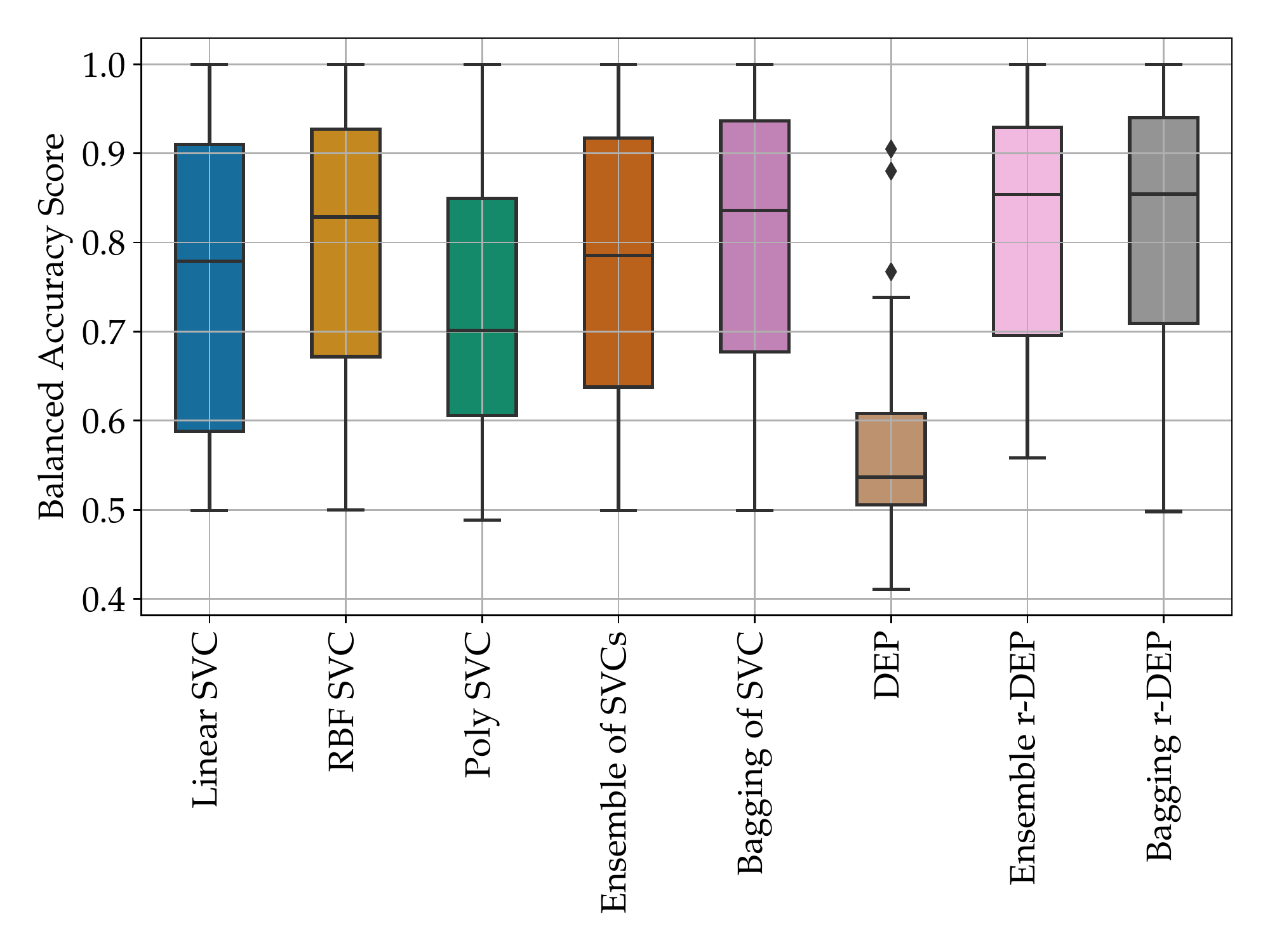}
    \caption{\new{Boxplot summarizing the average balanced accuracy scores provided on Table \ref{tab:accuracies}. In general, the ensemble and bagging r-DEP classifiers yielded largest balanced accuracy scores.}}
    \label{fig:BoxPlot}
\end{figure}

The outcome of the computational experiment is also summarized on the boxplot shown on Figure \ref{fig:BoxPlot}. The boxplot confirms that the ensemble and bagging r-DEP classifiers yielded, in general, the largest balanced accuracy scores. \new{This boxplot also reveals the poor performance of the DEP classifier which presupposes the positive samples are, in general, greater than or equal to the negative samples according to the component-wise ordering. In particular, the three points above the box of the DEP classifier corresponds to the average balanced accuracy score values 0.90, 0.88, and 0.77 obtained from the datasets Ionosphere, Breast Cancer Wisconsin, and Internet Advertisement, respectively. It turns out, however, that the ensemble and bagging r-DEP classifiers outperformed the original DEP model even in these three datasets. This remark confirms the important role of the transformations $\rho$ and $\sigma$ for successful applications of increasing lattice-based models.}   

\section{Concluding Remarks} \label{sec:Concluding} 

In analogy to Rosemblatt's perceptron, the morphological perceptron introduced by Ritter and Sussner can be applied for binary classification \cite{ritter96c}. In contrast to the traditional perceptron, however, the usual algebra is replaced by lattice-based operations in the morphological perceptron models. Specifically, the erosion-based and the dilation-based morphological perceptrons compute respectively an erosion $\varepsilon_{\vetm}$ and a dilation $\delta_{\vetw}$ given by \eqref{eq:ero_dil} followed by the application of the sign function. The erosion-based and dilation-based morfological perceptrons focus respectively on the positive and negative classes. A graceful balance between the two morphological perceptrons is provided by the dilation-erosion perceptron (DEP) classifier whose decision function given by \eqref{eq:DEP} is nothing but a linear combination of the an erosion $\varepsilon_{\vetm}$ and a dilation $\delta_{\vetw}$ \cite{araujo11}. 

In this paper, we propose to train a DEP classifier in two steps. First, based on the works of Charisopoulus and Maragos \cite{charisopoulos17}, the synaptic weights $\vetm$ and $\vetw$ of $\varepsilon_{\vetm}$ and $\delta_{\vetw}$ are determined by solving two independent convex-concave optimization problems given by \eqref{eq:objective}-\eqref{eq:ineq-positive} \cite{shen16}. Subsequently, the parameter $\beta$ is determined by minimizing the hinge loss given by \eqref{eq:hinge-loss}. 

Despite its elegant formulation, as a lattice-based model the DEP classifier presupposes that both feature and class spaces are partially ordered sets. The feature patterns, in particular, are ranked according to the component-wise ordering given by \eqref{eq:marginal}. Furthermore, the DEP classifier is an increasing operator. Therefore, it implicitly assumes a relationship between the orderings of features and classes. In many practical situations, however, the component-wise ordering is not appropriate for ranking features. Using results from multi-valued mathematical mophology, in this paper we introduced the reduced dilation-erosion perceptron (r-DEP) classifier. The r-DEP classifier corresponds to the r-increasing morphological operator derived from the DEP classifier $\phi$ by means of \eqref{eq:r-DEP-classifier} using a one-to-one correspondence $\sigma$ between the set of classes $\mathcal{C}$ and $\{-1,+1\}$ and a surjective mapping $\rho$ from the feature space $\mathbb{V}$ to $\bar{\R}^r$. Finding appropriate transformation mapping $\rho$ is the major challenge on the design of an r-DEP classifier.

Inspired by the supervised reduced ordering proposed by Velasco-Forero and Angulo \cite{velasco-forero11a}, we defined the transformation mapping $\rho$ using the decision functions of either an ensemble of SVCs with different kernels or a bagging of a base SVC trained using different samples of the original traning set. The source-codes of the ensemble and bagging r-DEP classifiers are available at \url{https://github.com/mevalle/r-DEP-Classifier}. Both ensemble and bagging r-DEP classifiers yielded the highest average of the balanced accuracy score among SVCs, their ensemble, and bagging of RBF SVCs, on 30 binary classification problems from the OpenML repository. Furthermore, paired Student's t-test with significance level at $99\%$ confirmed that the bagging r-DEP classifier outperformed the individual SVCs as well as their ensemble in our computational experiment. \new{The outcome of the computational experiment shows the potential application of the ensemble and bagging r-DEP classifier on practical pattern recognition problems including -- but not limited to -- credit card fraud detection or medical diagnosis. Moreover, although we only focused on binary classification, multi-class problems can be addressed using one-against-one or one-against-all strategies available, for instance, in the \texttt{scikit-learn API}}. 

In the future, we plan to investigate further the approaches used to determine the mapping $\rho$. We also intent to study in details the optimization problem used to train a r-DEP classifier.

\vspace{6pt} 




\funding{This research was funded by the S\~ao Paulo Research Foundation (FAPESP) under grant number 2019/02278-2 and the National Council for Scientific and Technological Development (CNPq) under grant number 310118/2017-4.}


\conflictsofinterest{The author declare no conflict of interest. The funders had no role in the design of the study; in the collection, analyses, or interpretation of data; in the writing of the manuscript, or in the decision to publish the results.} 

\reftitle{References}


\externalbibliography{yes}
\bibliography{references}



\end{document}